\documentclass[pdflatex,sn-mathphys]{sn-jnl}

\jyear{2023}%

\theoremstyle{thmstyleone}%
\theoremstyle{thmstyletwo}%

\theoremstyle{thmstylethree}%

\usepackage{multirow}
\usepackage{graphicx}
\usepackage{booktabs}
\usepackage{float}
\begin{document}

\title[Enhancing KPE from Long Scientific Documents using Graph Embeddings]{Enhancing Keyphrase Extraction from Long Scientific Documents using Graph Embeddings}


\author*[1]{\fnm{Roberto} \sur{Martínez-Cruz}}\email{rmcruz@comillas.edu}
\equalcont{These authors contributed equally to this work.}

\author[2]{\fnm{Debanjan} \sur{Mahata}}\email{dmahata@bloomberg.net}
\equalcont{These authors contributed equally to this work.}

\author[1]{\fnm{Alvaro J.} \sur{López-López}}\email{allopez@comillas.edu}
\author[1]{\fnm{José} \sur{Portela}}\email{jportela@comillas.edu}

\affil*[1]{\orgdiv{Institute for Research in Technology}, \orgname{ICAI School of Engineering, Comillas Pontifical University}, \orgaddress{\street{Madrid}, \country{Spain}}}

\affil[2]{\orgname{Bloomberg}, \orgaddress{\city{NYC}, \state{New York}, \country{USA}}}

\abstract{In this study, we investigate using graph neural network (GNN) representations to enhance contextualized representations of pre-trained language models (PLMs) for keyphrase extraction from lengthy documents. We show that augmenting a PLM with graph embeddings provides a more comprehensive semantic understanding of words in a document, particularly for long documents. We construct a co-occurrence graph of the text and embed it using a graph convolutional network (GCN) trained on the task of edge prediction. We propose a \textit{graph-enhanced} sequence tagging architecture that augments contextualized PLM embeddings with graph representations. Evaluating on benchmark datasets, we demonstrate that enhancing PLMs with graph embeddings outperforms state-of-the-art models on long documents, showing significant improvements in F1 scores across all the datasets. Our study highlights the potential of GNN representations as a complementary approach to improve PLM performance for keyphrase extraction from long documents.}

\keywords{keyphrase extraction, graph embeddings, long documents, pre-trained language models, natural language processing, deep learning}



\maketitle

\section{Introduction}
\label{introduction}

Keyphrase extraction (KPE) is a natural language processing task which involves identifying the most important phrases in a document that capture its main topics. These keyphrases provide a concise summary of the text and are useful for various downstream applications, such as document classification \cite{hulth2006study}, clustering \cite{hammouda2005corephrase}, summarization \cite{summarization-example-1, summarization-example-2}, indexing documents \cite{ie-example}, query expansion \cite{query-expansion-example}, interactive document retrieval \cite{jones1999phrasier} among others. The task is a form of extreme summarization, also known as keyphrasification \cite{keyphrasification}.

Most current algorithms for identifying keyphrases rely on using summarized versions of texts, such as scientific document abstracts, to determine the most important keyphrases. However, this approach has several drawbacks that limit its effectiveness. Firstly, in real-world scenarios, such as in industry applications, summaries may not always be available. As a result, the algorithm's performance may decline since it lacks training on longer texts. Secondly, important extractive keyphrases may be missing from the summaries. Additionally, crucial contextual information in the original text may not be reflected in the summaries, making it impossible to extract some of the keyphrases. As a result, this can significantly decrease the overall performance of the algorithm.

With the increasing reliance on digitization in diverse industry sectors such as legal, medical, scientific, and financial, where documents can span multiple pages, the need to identify keyphrases from these lengthy documents has become paramount. Extracting keyphrases can significantly improve search, enhance document understanding, save human annotation time, and facilitate knowledge discovery. Therefore, it is critical to develop a dedicated solution that can efficiently extract keyphrases from texts of any length, especially from long length documents. Such a solution will ensure that the benefits of KPE can be realized in a wide range of real-world scenarios, leading to improved performance and increased productivity.

KPE can be approached using unsupervised and supervised methods \cite{ccano2019keyphrasesurvey}. Unsupervised techniques typically use ranking algorithms to score phrases in a document based on their information content \cite{kesurvey2014}. One common approach is graph-based ranking \cite{graph-based-ke}, where a graph is built with words as nodes and edges representing relationships between linked words. The graph is then analyzed using node centrality measures to determine the importance of each phrase \cite{boudin2013comparison}. While graph-based ranking is a popular unsupervised method in this domain, the use of graph embeddings for learning word representations leveraging relationships between words in text documents, remains largely unexplored.

Supervised methods for Keyphrase Extraction (KPE) typically treat it as a token classification task. Tokens are represented numerically using features generated through manual feature engineering, such as term frequency, inverse document frequency, position in the title, and part-of-speech tags \cite{hasan2014automatic}. Alternatively, pre-trained word embeddings \cite{patel2019exploring}, such as word2vec \cite{word2vec}, GloVe \cite{glove}, or contextualized embeddings \cite{sahrawat2020keyphrase} calculated by a Pretrained Language Model (PLM) such as BERT \cite{bert}, can also be used. To optimize the use of these embeddings, a post-processing layer like a bidirectional long short-term memory (BiLSTM) \cite{bilstm} is often added to reproject the input embeddings. This is usually done into a lower-dimensional representation, while considering both previous and subsequent embeddings in the sequence.

If the embeddings are computed by a PLM, the model can be fine-tuned on the downstream task of KPE using a sequence tagging classification head. While supervised methods often achieve the best results, identifying keyphrases in long documents remains a challenge due to lack of techniques for embedding long dependencies between words. Such dependencies are crucial for understanding context in longer documents. Therefore, it is important to continue developing new methods that can effectively capture these long dependencies, leading to improved KPE performance in longer documents.

The current state-of-the-art results in KPE have been achieved using either a fine-tuned PLM \cite{kbir} or its raw embeddings as input for a trained model based on Bidirectional Long Short-Term Memory (BiLSTM) \cite{sahrawat2020keyphrase, park2020scientific}. Both approaches rely on contextualized representations of PLMs, which have the drawback of being limited to a maximum number of words. While this limitation is less significant when the text is short enough to be fully processed by the PLM such as scientific abstracts, it becomes increasingly problematic as the text gets longer.

To overcome the limitations of contextualized representations in KPE, we propose the use of graph embeddings applied to a co-occurrence graph. In this approach, each node represents a word, and the edges represent co-occurrence relationships between words within a sliding window. We use a GCN model \cite{gcn}, trained with the self-supervised task of link prediction, to embed each node. Previous approaches have attempted to utilize graph embeddings to improve the performance of the KPE task \cite{sasake,phraseformer}. However, none have explored the potential of using graph embeddings as a solution to address KPE in long documents, and none have leveraged the learned representations of GCN using self-supervised approach for this purpose.

Unlike supervised methods, our approach does not require annotated labels for training, which can be expensive and time-consuming to obtain. Instead, it leverages the natural co-occurrence patterns in the text to learn meaningful representations of words. This approach has previously achieved state-of-the-art results in downstream tasks, such as node similarity, demonstrating the effectiveness of GCN-based models for graph representation learning. 

By applying this approach to KPE, we aim to capture more comprehensive information about the document's content, including long-term dependencies between words, and enhance the performance of the model on longer texts. In our proposed model as described in Section \ref{methodology} we supplement the representations from a PLM with the representations from the learnt graph embedding. With our proposed methodology we hypothesize the following advantages:
\begin{itemize}
    \item By representing the text as a co-occurrence graph and applying graph embeddings, we can capture relationships between words in a non-sequential manner. This approach has the advantage of embedding distant but related words closer in the semantic space if they have co-occurred or are related through co-occurring words. Unlike sequential models that process words in a fixed order, graph embeddings can capture more complex relationships between words in the context of the input document and reflect the underlying semantic structure of the text. This makes them a more suitable representation for embedding long-term relationships between words, which are often crucial for understanding the context of longer texts. 

    \item The graph-based representation captures all relationships in the text without being constrained by the input word limit. As a result, the graph embedding provides a semantic understanding of a word's usage throughout the entire text, eliminating any length restrictions that may exist. This enables the numerical representation of words to be extended to encompass the entirety of the text.
\end{itemize}

In this work, we propose a \textit{graph-enhanced sequence tagging} approach (Section \ref{seq-tag}) to address the limitations of existing PLM-based methods for keyphrase extraction on longer texts by augmenting them with graph embeddings. By leveraging the non-sequential nature of graph representations, we aim to capture more semantically rich information about the document's content and enhance the performance of the model on longer texts. The major contributions of the work as proposed in this paper are: 
\begin{itemize}
    \item We introduce a novel framework (Section \ref{methodology}) that combines graph embeddings and PLMs for keyphrase extraction on longer texts. To the best of our knowledge, this is the first study that attempts to do it in the context of long documents. 
    \item We demonstrate the effectiveness of our proposed approach on five popular benchmark datasets (Section \ref{experiments}) and show that it outperforms state-of-the-art methods based on PLMs. 
    \item We provide a comprehensive analysis of the results and highlight the strengths and limitations of our approach, as well as opportunities for future research.
\end{itemize}

Next, we provide a brief review of the related work in the field (Section \ref{related}), focusing on the topic of our paper. We then present our proposed methodology in detail (Section \ref{methodology}), followed by a description of the experiments we conducted (Section \ref{experiments}) and the results we obtained (Section \ref{results}). In addition, we discuss the limitations of our approach and suggest future directions for research in this area (Section \ref{conclusion}).
\section{Related Work}
\label{related}

\subsection{Keyphrase Extraction}

There are two main approaches for Keyphrase Extraction (KPE): supervised and unsupervised. Unsupervised methods typically rely on graph-based approaches, while supervised methods do not commonly use them. The unsupervised approach involves several phases, starting with the extraction of candidate words or phrases from the text using heuristics such as POS patterns for words or n-grams. The next step is to score and rank the extracted phrases according to their importance. Graph-based ranking techniques are popular in this last phase, where centrality measures are applied to the co-occurrence graph of the text. These centrality measures are mostly derived from Google’s PageRank \cite{pagerank} algorithm and are combined with other relevant measurements based on clustering, tf-idf scores, extraction of specific lexical patterns, and word co-occurrence measures. Examples of this methodology include TextRank \cite{textrank} and TopicRank \cite{topicrank}. More recent methods have incorporated word embeddings to improve the accuracy of the ranking phase \cite{wang2014corpus, mahata2018key2vec, 
mahata2018theme, bennani2018simple}. Unsupervised methods have proven to be effective when working with unannotated data. However, when labeled data is available, supervised methods tend to perform better and can be tailored to the specific linguistic and contextual characteristics of the topic.



Initially, supervised approaches for KPE relied on manually-engineered features extracted from the text, such as term frequencies \cite{hulth2003}, syntactic properties \cite{kim-kan-2009-examining}, or location information \cite{nguyen-kan-2007}, to represent each word in the sequence. These features were then fed into a classifier to independently classify each candidate phrase as a keyphrase or not. However, it wasn't until the work of \cite{Gollapalli-Li-Yang-2017} that KPE was approached as a sequence labeling task, using a conditional random field (CRF) to take advantage of the sequential nature of the data. In an improved version of the sequence labeling approach, \cite{alzaidy-2019} replaced manual feature engineering with pre-trained GloVe embeddings to represent each word in the sequence. They used a bidirectional long short-term memory (BiLSTM) CRF layer to exploit this representation. The BiLSTM layer updated the representation of the input sequence of words with contextual information, while the CRF modeled the dependencies between the sequence of classifications produced by the BiLSTM outputs.

The transformer architecture \cite{attention-is-all-you-need} has greatly improved results for various natural language processing tasks, including keyphrase extraction in works such as TransKP \cite{transkp} and TNT-KID \cite{tnt-kid}. It is effective at embedding words in a sequence, with representations that depend not only on the word itself but also on its context. This has led to the development of PLMs that specialize in providing contextualized embeddings of words in a sentence. In combination with a trained BiLSTM-CRF layer, these embeddings have outperformed all previous models, as demonstrated in the work of \cite{sahrawat2020keyphrase}. Some works, such as \cite{kbir}, have tried to further enhance the representation of these embeddings by designing self-supervised objectives specifically for keyphrase extraction tasks, resulting in KBIR\footnote{https://huggingface.co/bloomberg/KBIR} (Keyphrase Boundary Infilling and Replacement), a PLM that achieved state-of-the-art results. 

The sequence labeling approach has a limitation in that it considers only a limited number of words as context. This limitation can result in poor performance on longer texts where the number of words exceeds the maximum input limit. Prior works such as Phrasephormer \cite{phraseformer} and SaSAKE \cite{sasake} have integrated graph embeddings into their architecture, however, none have leveraged them to enhance keyphrase identification in long documents. SaSAKE did not utilize any self-supervised learned representation of the graph, while Phrasephormer did not incorporate GNNs and instead relied on older graph embedding methods that were found to be less effective than GNN architectures in generating rich representations. 

It is worth noting that other solutions in the field of Keyphrase Generation (KPG), that addresses the problem as a text generation task to simultaneously extract and abstract keyphrases, have incorporated GNNs to tackle the task, as seen in \cite{divgraphpointer} and \cite{heter-graph-kpg}. However, these solutions differ from our approach as they focus on text generation instead of sequence tagging and do not utilize self-supervised learned representations of the graph.

\subsection{Graph Embeddings}

Graphs are commonly employed to represent various types of networks, and the challenge lies in developing a low-dimensional numerical representation that captures the context of a node in the network. However, unlike text and images, graphs lack a pre-established logical order, making the task particularly challenging. Early approaches to address this issue were based on pre-trained word embeddings such as word2vec \cite{word2vec}, which used random walks in the network to predict the next node in the walk given the one-hot encoding of the current node. DeepWalk \cite{deepwalk} and Node2Vec \cite{node2vec} adopted this approach, with the latter introducing bias through certain design parameters. However, these methods had limitations such as not utilizing node and edge features and being unable to incorporate new nodes after training. To address these shortcomings, Graph Convolutional Network (GCN) \cite{gcn} was developed as a neural network specifically designed for solving graph problems. GCN embeddings integrate node and edge features with information from surrounding nodes in the network into a single representation and can be easily updated with new nodes. Other variations of GCN, such as SageGraph \cite{sagegraph} and Graph Attention Networks (GAT) \cite{gat}, have since been developed to enhance the performance of the original architecture, with GAT introducing attention mechanisms to improve the model's focus on the relevant nodes. 

Our research is inspired by the concept of link prediction, which has been explored in SageGraph, as well as earlier studies that employ this task to produce a low-dimensional representation of nodes in the graph. For instance, \cite{zhang2022link} investigates two types of structural contexts, namely context nodes obtained from random walks and context subgraphs, to create node embeddings customized for link prediction. Similarly, \cite{saxena2021nodesim} introduces a network embedding technique, called NodeSim, that captures both node similarities and community structure to learn the low-dimensional representation of the network. Building on these insights, we propose a self-supervised method that utilizes link prediction to train a GCN, which can effectively embed the nodes in the graph. To the best of our knowledge, our approach represents the first attempt to apply this methodology in the NLP domain.

\subsection{NLP and KPE from Long Documents}

Dealing with long sequences of text can significantly increase the complexity of any NLP task, not only due to the added length but also because long-term dependencies are difficult to embed effectively. Most PLMs have limitations on the maximum number of words they can handle, restricting the context. Additionally, the transformer's multi-attention architecture presents challenges when dealing with a large number of words. Two PLMs have emerged as promising solutions that address these issues by incorporating novel variations in their modeling: Big Bird \cite{bigbird} and Longformer \cite{longformer}. 

Big Bird, introduced by Google, is specifically designed for processing long documents and employs a sparse attention mechanism that allows the model to attend to a subset of tokens at each layer, reducing the computational cost. It also employs a global attention mechanism to attend to the entire document while maintaining sparsity. The original model can handle up to 2,048 input tokens, while a variation called BigBird-PEGASUS can handle up to 4,096. Longformer, a transformer-based language model is designed to process long documents with a sequence length of up to 4,096 tokens, uses a ``sliding window attention" mechanism to attend to tokens across the entire sequence while maintaining a quadratic computation cost. These two models have shown promising results on various NLP tasks that involve processing long documents.

Numerous task-specific solutions have been proposed to tackle the challenge of processing long documents. For instance, in text classification, \cite{classification-of-long-documents} uses transformers to efficiently process lengthy documents. In document understanding, \cite{understanding-long-documents} improves attention mechanisms for long texts. \cite{grail-etal-2021-globalizing} extends BERT-based architectures for summarizing long documents. \cite{bertal} utilizes a combination of a Transformer and multi-channel LSTM to handle long texts without requiring re-pretraining. In document matching, \cite{beyond-512} uses a multi-depth transformer-based encoder to handle long-form document matching tasks.

Although Keyphrase Extraction (KPE) is a crucial task in natural language processing, only a few works have attempted to address the challenge of extracting keyphrases from lengthy documents. One example of such work is \cite{mahata2022ldkp}, which released two large datasets containing fully extracted text and metadata. This study reported on the performance of various unsupervised and supervised algorithms for keyphrase extraction. Another noteworthy example is \cite{query-based-kpe}, which proposes a system that chunks documents while maintaining a global context as a query for relevant keyphrase extraction. To estimate the probability of a given text span forming a keyphrase, the system employs a pre-trained BERT model. Interestingly, the results demonstrate that a shorter context with a query outperforms a longer context without a query. To enhance the generation of keyphrases, \cite{garg2022keyphrase} investigated the inclusion of information beyond the title and abstract as input in the field of KPG. The observed improvement in results through their approach demonstrates that the model must not solely rely on the summary provided by the title and abstract to predict high-quality keyphrases. These studies shed light on the importance of developing effective methods for keyphrase extraction from lengthy documents and offer promising directions for future research.

Our approach is centered on providing the model with a comprehensive representation of the long-term dependencies between words, an aspect that has received limited attention in prior research. We posit that this information can be effectively incorporated through the graph embedding, enabling the model to overcome its contextual limitations and leverage the embedding to expand its understanding of the input document. To the best of our knowledge, no other study has harnessed graph embeddings for this purpose, and we believe our approach has the potential to significantly enrich the state-of-the-art for KPE from lengthy documents.

\section{Methods}
\label{methodology}\
In this section, we formulate Keyphrase Extraction (KPE) as a sequence tagging task, and provide details on the architecture used to embed the nodes of the co-occurrence graph. We also describe the calculation of contextual embeddings of the words and explain how both representations are utilized by a sequence tagging model to identify keyphrases in a given text.

\subsection{Problem Formulation}
The task is to classify each word $w_{t}$ in a given input sequence of words $d = \{w_1, w_2, ..., w_n\}$ into one of three possible classes: $Y = {K_{B}, K_{I} , O}$. In this context, $K_{B}$ represents the beginning word of a keyphrase, $K_{I}$ indicates a word is inside a keyphrase, and $O$ denotes that $w_t$ is not part of a keyphrase. This process is commonly referred to as BIO tagging scheme in the sequence tagging literature.

\subsection{Graph Embedding}
In this section, we provide a detailed description of the process of converting a document into a co-occurrence graph and how a Graph Convolutional Network (GCN) is trained for self-supervised edge prediction to obtain node embeddings from the words in the document.

\subsubsection{Co-occurrence Graph}
\begin{figure}[H]
    \centering
    \includegraphics[scale=0.3]{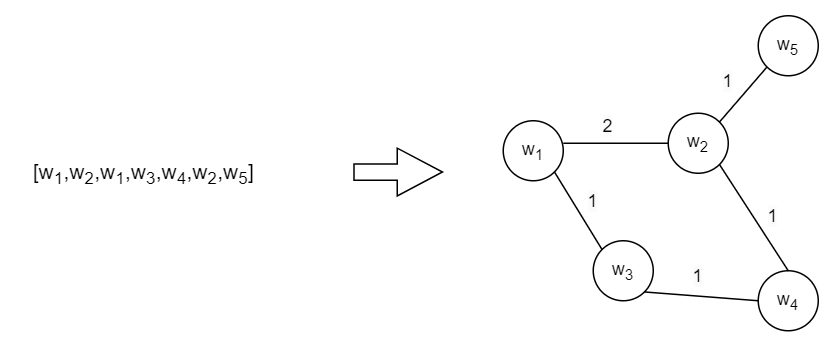}
    \vspace{2mm}
    \caption{Co-occurrence graph construction with a sliding window size ($s_d$) = 2}
    \label{fig:coocurency_graph}
\end{figure}

To transform the input sequence of words $d = \{w_1, w_2, ..., w_n\}$ into an undirected graph, we create a graph $G = (V,E)$, where $V$ represents the vertices of the graph, and each unique word in $d$ is represented as a vertex. The edges of the graph, denoted as $E$, indicate co-occurrences between words within a sliding window of size $s_d$. The weight of each edge is determined by the number of times the words it connects co-occur within sliding windows. Figure \ref{fig:coocurency_graph} shows an example of this transformation using a sliding window size of 2.

\subsubsection{Graph Convolutional Network}
In this section, we present the forward propagation algorithm, as shown in Algorithm \ref{alg:alg1}, that is utilized for embedding the nodes in the graph $G = (V,E)$, assuming that the model parameters have already been trained. Specifically, we assume that the set of weight matrices $W_k$, $\forall k \in \{1, ..., K\}$ is fixed, and these matrices are employed to propagate information between different layers of the model.

\begin{algorithm}[H]
    \caption{GCN embedding generation algorithm}\label{alg:alg1}
    \textbf{Input :} Graph $G(V, E)$; input features $\{x_v, \forall v \in V\}$; edge weights $\{ew_{u-v}, \forall u,v \in V\}$; depth $K$; embedding layer $EMBED$, weight matrices $W_k, \forall k \in \{1, ..., K\}$; non-linearity $\sigma$; differentiable aggregator functions $AGGREGATE_k, \forall k \in \{1, ..., K\}$; neighborhood function $N : v \to 2^V$\\
    \textbf{Output :} Vector representations $z_v$ for all $v \in V$
    \begin{algorithmic}[1]
        \State $h^0_v \gets x_v, \forall v \in V$
        \For{$k \gets 1$ \textbf{to} $K$}
        \For{$v \in V$}
        \State $h_{N(v)} \gets \{EMBED(u), \forall u \in N(v)\})$
        \State $h^k_{N(v)} \gets AGGREGATE_k(\{ew_{u-v}*h^{k-1}_u, \forall u \in N(v)\})$
        \State $h^k_v \gets \sigma(W_k \cdot \text{CONCAT}(h^{k-1}_v, h^k_{N(v)}))$
        \EndFor
        \State $h^k_v \gets \frac{h^k_v}{\sqrt{k\left\|h^k_v\right\|^2}}, \forall v \in V$
        \EndFor
        \State $z_v \gets h^K_v, \forall v \in V$
    \end{algorithmic}
\end{algorithm}

\begin{figure}[H]
    \centering
    \includegraphics[scale=0.8]{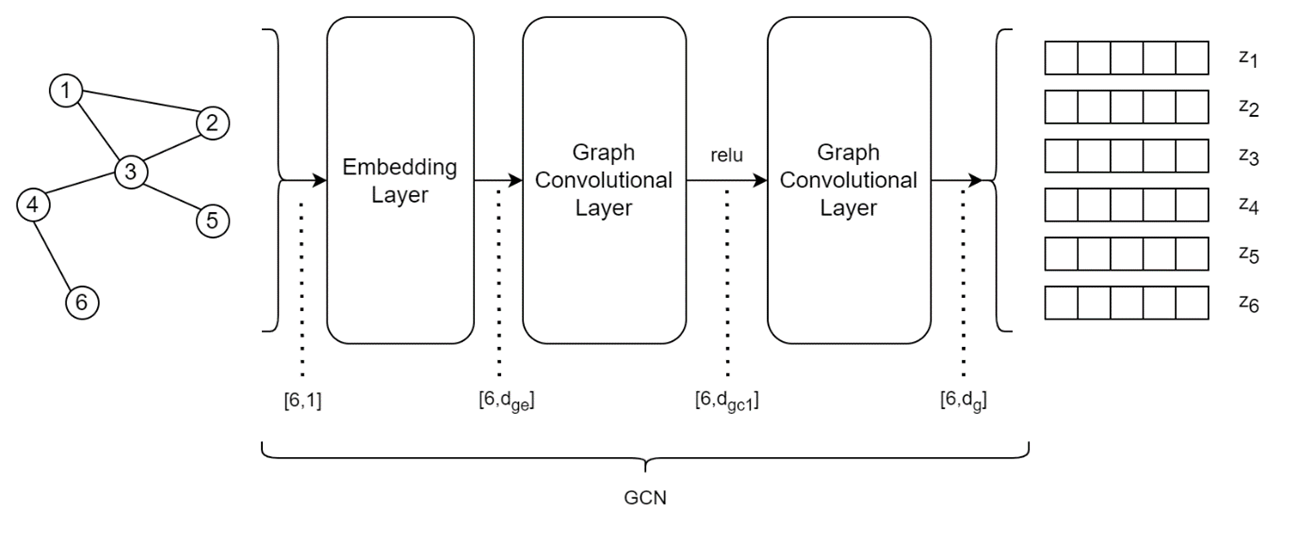}
    \caption{A schematic example of node embedding using GCN.}
    \label{fig:gcn}
\end{figure}

In the scope of this work, we consider all edges without sampling, and the edge weight $e_w$ for nodes that have no connection is set to 0. After applying Algorithm \ref{alg:alg1} to all the nodes, the resulting matrix $Z = \{z_1, z_2, \dots, z_n\}$ represents the $d_g$-dimensional embedding of the $n^{th}$ node. A schematic example of the process is illustrated in Figure \ref{fig:gcn}.


\subsubsection{Self-supervised Edge Prediction}

Utilizing the node representation provided by the GCN, we calculate $E_{p_{i,j}}$, which denotes the probability of the $i^{th}$ and $j^{th}$ nodes being connected. To construct the training set for this task, we utilize negative sampling of the existing edges in $G$. Specifically, for each existing link in the network labeled as 1, we generate $n_{ns}$ non-existent edges and label them as 0.


The training data comprises two sets: $X_g = \{e_1, e_2, \dots, e_t\}$ and $Y_g = \{y_1, y_2, \dots, y_t\}$. Each element $e_t$ in $X_g$ represents the $t^{th}$ edge in the dataset and is composed of two nodes, $n_{t_{1}}$ and $n_{t_{2}}$, from the graph. The corresponding element $y_t$ in $Y_g$ is a boolean value that indicates whether the nodes forming $e_t$ are connected.

\begin{figure}[H]
    \centering
    \includegraphics[scale=0.8]{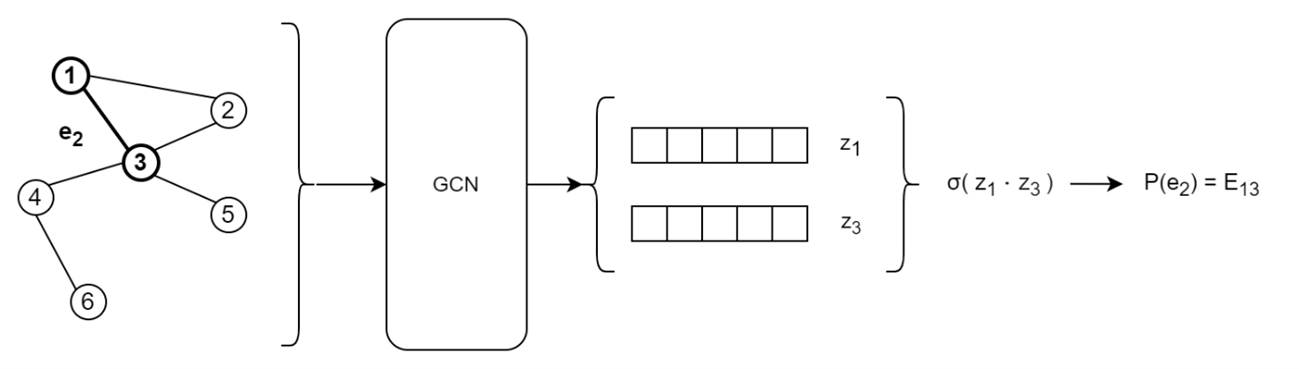}
    \caption{GCN edge prediction training - Schematic example}
    \label{fig:edge_prediction}
\end{figure}

To estimate the probability $E_{p_{i,j}}$ that nodes $i$ and $j$ are connected, the dot product of their respective embeddings $z_i$ and $z_j$ is calculated. To compute the probabilities for all possible node pairs, a probability matrix $E_p$ of size $n_{nodes} \times n_{nodes}$ is constructed, where each entry $(i,j)$ represents the probability that nodes $i$ and $j$ are connected. $E_p$ is obtained by computing the matrix multiplication of $Z$ with its transpose $Z^T$, as outlined in Algorithm \ref{alg:alg1}. Figure \ref{fig:edge_prediction} provides a schematic example of this calculation.

To train the network using gradient descent, the binary cross-entropy loss function is utilized, as shown in Equation \ref{eq:eq1}. Here, $n_g$ represents the length of the training dataset, and $\sigma$ denotes the sigmoid function.

\begin{equation}
    \label{eq:eq1}
    L_g = -\frac{1}{n_g} \sum\limits_{i=1}^{x_g} (y_i * \log(\sigma(z_{n_{i1}} * z_{n_{i2}})) + (1 - y_i) * \log(1 - \sigma(z_{n_{i1}} * z_{n_{i2}})))
\end{equation}

\subsection{Contextual Embedding}
\label{contextual-embedding}



This component is responsible for utilizing a pre-trained language model (PLM) to convert the input sequence of words $d = {w_1, w_2, \dots, w_n}$ into a sequence of contextual embeddings $h = {h_1, h_2, \dots, h_n}$, where $h_{c_n}$ denotes the $d_c$-dimensional embedding for the $n^{th}$ word.

\begin{figure}[H]
    \centering
    \includegraphics[scale=0.75]{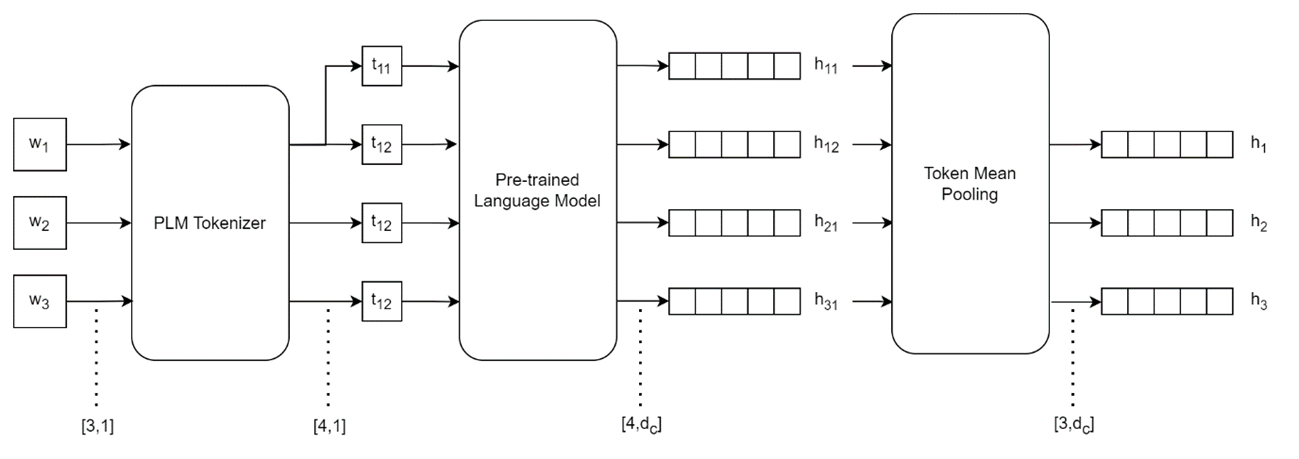}
    \caption{Schematic diagram for calculating contextual embeddings.}
    \label{fig:contextualized_embedding}.
\end{figure}

Since KPE is treated as a sequence tagging problem, specialized language models are employed for this task. These PLMs, such as BERT or KBIR, utilize the transformers encoder as the base model and pre-train it on a large corpus of text using a self-supervised objective.
These models tokenize the words into sub-words before processing. To aggregate the results and obtain a unique embedding for each word, mean-word pooling is used, where the embedding of a word is the mean for each dimension of its sub-words' embeddings. Figure \ref{fig:contextualized_embedding} provides a schematic example.

\subsection{Graph Enhanced Sequence Tagger}
\label{seq-tag}

\begin{figure}[]
    \centering
    \includegraphics[scale=1]{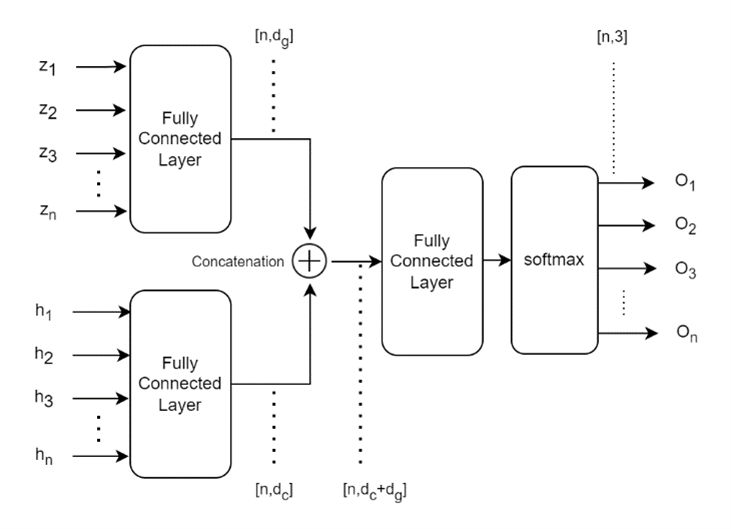}
    \caption{A schematic diagram of the graph-enhanced sequence tagger.}
    \label{fig:fc_model}
\end{figure}

After undergoing the aforementioned processes, the input words $w$ are transformed into two sets of embeddings: a sequence of graph embeddings $z$ and a sequence of contextual embeddings $h$. To further enhance their fusion, these embeddings are individually passed through separate fully connected layers before being concatenated into a unified numerical representation for each word. This resulting sequence of embeddings is then fed to a fully connected layer followed by a softmax function to obtain a sequence of probabilities for each BIO tag. This architecture is utilized for fine-tuning the PLM. For a visual representation of the process, please refer to the schematic example in Figure \ref{fig:fc_model}.

In cases where the input sequence is larger than the PLM's maximum allowable number of words, a sliding window approach is employed. The window size is set to the upper limit of words for the PLM, which is also used as the stride. This enables the model to process the entire sequence while staying within the constraints of the PLM's input size limitations. In the rest of the paper we refer to this sequence tagging model as the \textit{Graph Enhanced Sequence Tagger}.

\section{Experiments}
\label{experiments}

In this section, we describe the experimental settings and report the performance of our proposed model on a series of experiments conducted to investigate its effectiveness on multiple benchmark datasets for KPE. We also discuss the implications of our findings and draw conclusions based on the observations from the obtained results.


\subsection{Datasets}
We evaluated our approach using four publicly available datasets consisting of full-length long documents from two domains: scientific and news. Specifically, we used the SemEval-2010 \cite{kim-2010}, LDKP3K \cite{mahata2022ldkp}, NUS \cite{nguyen-kan-2007}, and DUC-2001 \cite{duc2001} datasets. To assess the feasibility and effectiveness of our method on short abstracts, we also report performance on the Inspec dataset \cite{hulth2003}. As discussed in Section \ref{methodology}, we formulate KPE as a sequence tagging problem and focus on extractive keyphrases that are present in the text. We label each word in the samples using the B-I-O tagging scheme. Any text input longer than 512 tokens/words are considered as a long document. Brief descriptions of each dataset are provided below.

\begin{enumerate}
    \item The \textbf{SemEval-2010} dataset\footnote{\url{https://huggingface.co/datasets/midas/semeval2010}} is composed of 284 full-length ACM articles, which have been divided into train, trial, and test sets, containing 144, 40, and 100 articles, respectively. 
    \item \textbf{LDKP3K}\footnote{\url{https://huggingface.co/datasets/midas/ldkp3k}} is a collection of roughly 100,000 keyphrase-tagged long documents that were generated by mapping the KP20K corpus to S2ORC \cite{s2orc-corpus}. For our experiments, we utilized the smaller version of this dataset, which includes 20,000 training samples, 3,413 validation samples, and 3,339 test samples.
    \item The \textbf{NUS} dataset\footnote{\url{https://huggingface.co/datasets/midas/nus}} comprises 211 scientific documents that were extracted and chosen utilizing the Google SOAP API. These documents were manually annotated and are solely used for evaluation, as the dataset only includes a test split.
    \item The \textbf{DUC-2001} dataset\footnote{\url{https://huggingface.co/datasets/midas/duc2001}} is composed of 308 news articles gathered from TREC-9, each of which is manually annotated with controlled keyphrases for evaluation purposes. Only a test split is included in the dataset.
    \item \textbf{Inspec}\footnote{\url{https://huggingface.co/datasets/midas/inspec}} includes abstracts from 2,000 scientific articles, which have been split into three sets for training, validation, and testing purposes. The training set consists of 1,000 abstracts, while the validation and test sets contain 500 abstracts each.
\end{enumerate}

Detailed statistics for these datasets can be found in Table \ref{tab:dataset_stats}.

\begin{table}[]
\centering
\begin{tabular}{@{}lllll@{}}
\toprule
\textbf{Dataset} & \textbf{Size} & \textbf{Long Doc} & \textbf{Domain} & \textbf{Avg \# Words} \\ \midrule
Inspec           & 2k            & No                & Scientific      & 130.57                \\ \midrule
DUC-2001          & 0.308k        & Yes               & News            & 740                   \\ \midrule
SemEval-2010      & 0.24k         & Yes               & Scientific      & 7,434.52              \\ \midrule
LDKP3K           & 100k          & Yes               & Scientific      & 6,027.10              \\ \midrule
NUS              & 0.21k         & Yes               & Scientific      & 7,644.43              \\ \bottomrule
\end{tabular}
\vspace{3mm}
\caption{Statistics of the datasets used in our experiments}
\label{tab:dataset_stats}
\end{table}

 

\subsection{Evaluation Metrics}

We utilized $F1@K$ as our evaluation metric \cite{kim-2010}. Equations \ref{precision}, \ref{recall}, and \ref{f1} demonstrate how to calculate $F1@K$. Before evaluation, we lowercased, stemmed, and removed punctuation from the ground truth and predicted keyphrases and used exact matching. Let $Y$ denote the ground truth keyphrases, and $\bar{Y} = (\bar{y_1},\bar{y_2}, \dots, \bar{y_m})$. Then, we can define the metrics as follows:

\begin{equation}
    Precision@k =\frac{\vert Y \cap Y_k \vert}{\min\{\vert Y_k \vert, k\}}
\label{precision}
\end{equation}
\vspace*{1mm}

\begin{equation}
    Recall@k = \frac{\vert Y \cap Y_k \vert}{\vert Y \vert}
\label{recall}
\end{equation}
\vspace*{1mm}

\begin{equation}
    F1@k = \frac{2*Precision@k*Recall@k}{Precision@k + Recall@k}
\label{f1}
\end{equation}
\vspace*{3mm}

Where $\bar{Y}_k$ denotes the top $k$ elements of the set $\bar{Y}$. In our case, $k$ is equal to $K$, which denotes the total number of predicted keyphrases.

\subsection{Setup}
The primary aim of the experiments is to investigate whether the incorporation of graph embeddings can improve the quality of word representations in pre-trained language models (PLMs) by providing additional contextual information. The research employs various PLMs, including BERT\footnote{\url{https://huggingface.co/bert-base-uncased}} \cite{bert}, SciBERT\footnote{\url{https://huggingface.co/allenai/scibert_scivocab_uncased}} \cite{scibert} (a specialized variant of BERT for scientific content), DistilBERT\footnote{\url{https://huggingface.co/distilbert-base-uncased}} (a distilled version of BERT), KBIR\footnote{\url{https://huggingface.co/bloomberg/KBIR}} \cite{kbir} (a RoBERTa-based model that achieves state-of-the-art performance in the KPE task), and Longformer\footnote{\url{https://huggingface.co/allenai/longformer-base-4096}} \cite{longformer} (a transformer-based model specifically designed for efficient handling of long input sequences). Pertinent model details are presented in Table \ref{tab:model-specs}.

\begin{table}[]
\centering
\resizebox{\textwidth}{!}{%
\begin{tabular}{@{}llll@{}}
\toprule
\textbf{Model} & \textbf{Domain}                                                           & \textbf{\begin{tabular}[c]{@{}l@{}}Maximum Input\\ Tokens\end{tabular}} & \textbf{\#Parameters} \\ \midrule
DistilBERT     & MultiDomain                                                                           & 512                                                                     & 66.4M                \\ \midrule
BERT           & MultiDomain                                                                           & 512                                                                     & 109.5M               \\ \midrule
SciBERT        & Scientific                                                                            & 512                                                                     & 109.9M               \\ \midrule
KBIR           & \begin{tabular}[c]{@{}l@{}}MultiDomain -\\ Specialized in KPE task\end{tabular}       & 512                                                                     & 355.4M               \\ \midrule
LongFormer     & \begin{tabular}[c]{@{}l@{}}MultiDomain -\\ Specialized in Long Documents\end{tabular} & 4096                                                                    & 148.7M               \\ \bottomrule
\end{tabular}%
}
\vspace{3mm}
\caption{Details of models used in our experiments.}
\label{tab:model-specs}
\end{table}
 

Throughout our experiments, we conducted fine-tuning on all PLMs using two approaches: (a) popularly used token classification approach\footnote{https://huggingface.co/tasks/token-classification} for sequence tagging, (b) \textit{graph-enhanced} sequence tagging approach as explained in Section \ref{seq-tag}. Our aim was to assess the ability of the PLMs to assimilate the additional information derived from the graph embeddings during training and to determine whether the inclusion of such embeddings could enhance the fine-tuning process leading to improvement in performance.

For instance, we utilized SciBERT to explore the possibility of enhancing the PLM's representation by incorporating graph embeddings, despite its customization to a particular domain. Through this approach, we aimed to examine the potential of graph embeddings in providing supplementary information beyond the domain-specific knowledge.

Furthermore, we employed Longformer, a transformer model with an expanded token capacity (4,096 tokens), to investigate the effectiveness of our approach. Specifically, we aimed to assess whether the incorporation of graph information could improve the quality of representation, even in the presence of a robust contextual understanding of long sequences in the longformer model.


We evaluated the effectiveness of the models on both long and short documents, demonstrating the viability and effectiveness of our strategy for both forms of input length. Specifically, for short documents, we trained and evaluated our models exclusively on the Inspec corpus. For long documents, we trained our models on two distinct datasets, SemEval-2010 and LDKP3K, and assessed their performance on the test splits of all the datasets. Notably, the training set of SemEval-2010 contains only 144 documents, which presents a challenging few-shot learning scenario. In contrast, LDKP3K includes a larger training set of 20,000 samples, enabling us to validate the efficacy of our strategy across both small and large training data.
 
As the Inspec dataset comprises short documents that can be accommodated within BERT's context limit, we only employed BERT to evaluate the performance of our models. To establish a baseline for comparison, we trained and tested all the proposed architectures without incorporating graph embeddings as input.



To construct the co-occurrence matrix for each input document, we employed a window size of 4. For the graph embedding calculation, we created the link prediction dataset with a negative sampling ratio of 5. Subsequently, we trained a dedicated GCN model for each training dataset for 5 epochs with the mean as the aggregation function. The embedding dimension was set to 192, as were the dimensions of the first and second graph convolutional layers. To evaluate the model's performance, we used the AUC-ROC score and selected the best-performing model for node embedding calculation.
 
The sequence tagging models were trained using mini-batching with a batch size of 10. The models underwent 100 epochs of training, utilizing the AdamW algorithm with a learning rate of 5e-10, a patience value of 5, and an annealing factor of 0.5. During fine-tuning, the baseline models were trained solely with the classifier head. In contrast, for our proposed strategy of a \textit{Graph Enhanced Sequence Tagger}, we integrated the graph embeddings and fused their representations with that of the PLM, as depicted in Fig \ref{fig:fc_model}.

\section{Results}
\label{results}
In this section, we present the results of our experiments. For each model variant, we report the average score on the test set for three models trained with different random seeds. We observed that incorporating graph embeddings into the representation consistently improved the performance of the models across all settings. The improvement was particularly notable for longer documents. To name our models, we follow the convention of indicating the PLM embedding used in the first part of the name, and appending the term ``Graph Enhanced" for the \textit{Graph Enhanced Sequence Tagger} using the same PLM.

\subsection{Long Documents}


We conducted experiments to assess the effectiveness of incorporating graph embeddings in contextualized word representations for lengthy documents that exceed the PLM's input length limitation. The models were trained on SemEval-2010 and LDKP3K datasets, and the results of these experiments are presented in Table \ref{tab:semeval-results} (for models trained on SemEval-2010 train data) and Table \ref{tab:ldkp3k-results} (for models trained on LDKP3K train data), respectively.

Our results demonstrate that graph embeddings enhance the representation in all settings, indicating that they provide crucial contextual information to the sequence tagging model, which is crucial for the KPE task. Notably, the performance improvement is particularly significant for domain-specific models like SciBERT. We postulate that there is a mutually beneficial relationship between leveraging domain-specific pre-training and utilizing graph embedding information. Nonetheless, all models benefit from incorporating graph embeddings, highlighting the importance of our approach in capturing essential information for KPE from lengthy sequences that cannot be obtained through domain or task-specific pre-training of PLMs.

\begin{table}[]
\centering
\resizebox{\textwidth}{!}{%
\begin{tabular}{@{}|c|c|c|c|c|c|c|c|c|@{}}
\toprule
\textbf{Model} &
  \textbf{SemEval-2010} &
  \textbf{\begin{tabular}[c]{@{}c@{}}\% \\ improve\end{tabular}} &
  \textbf{DUC-2001} &
  \textbf{\begin{tabular}[c]{@{}c@{}}\% \\ improve\end{tabular}} &
  \textbf{NUS} &
  \textbf{\begin{tabular}[c]{@{}c@{}}\% \\ improve\end{tabular}} &
  \textbf{LDKP3K} &
  \textbf{\begin{tabular}[c]{@{}c@{}}\% \\ improve\end{tabular}} \\ \midrule
\textbf{DistilBERT} &
  0.228 ± 0.011 &
  \multirow{2}{*}{3.07\%} &
  0.034 ± 0.012 &
  \multirow{2}{*}{67.65\%} &
  0.170 ± 0.005 &
  \multirow{2}{*}{1.76\%} &
  0.108 ± 0.006 &
  \multirow{2}{*}{9.26\%} \\ \cmidrule(r){1-2} \cmidrule(lr){4-4} \cmidrule(lr){6-6} \cmidrule(lr){8-8}
\textbf{\begin{tabular}[c]{@{}c@{}}Graph Enhanced \\ DistiBERT\end{tabular}} &
  0.235 ± 0.004 &
   &
  0.057 ± 0.003 &
   &
  0.173 ± 0.006 &
   &
  0.118 ± 0.005 &
   \\ \midrule
\textbf{BERT} &
  0.240 ± 0.004 &
  \multirow{2}{*}{2.08\%} &
  0.078 ± 0.005 &
  \multirow{2}{*}{8.97\%} &
  0.173 ± 0.004 &
  \multirow{2}{*}{2.89\%} &
  0.118 ± 0.001 &
  \multirow{2}{*}{3.39\%} \\ \cmidrule(r){1-2} \cmidrule(lr){4-4} \cmidrule(lr){6-6} \cmidrule(lr){8-8}
\textbf{\begin{tabular}[c]{@{}c@{}}Graph Enhanced \\ BERT\end{tabular}} &
  0.245 ± 0.001 &
   &
  \textbf{0.085} ± 0.014 &
   &
  0.178 ± 0.006 &
   &
  0.122 ± 0.003 &
   \\ \midrule
\textbf{SciBERT} &
  0.236 ± 0.007 &
  \multirow{2}{*}{4.24\%} &
  0.029 ± 0.006 &
  \multirow{2}{*}{13.79\%} &
  0.178 ± 0.010 &
  \multirow{2}{*}{8.99\%} &
  0.133 ± 0.012 &
  \multirow{2}{*}{10.53\%} \\ \cmidrule(r){1-2} \cmidrule(lr){4-4} \cmidrule(lr){6-6} \cmidrule(lr){8-8}
\textbf{\begin{tabular}[c]{@{}c@{}}Graph Enhanced \\ SciBERT\end{tabular}} &
  0.246 ± 0.014 &
   &
  0.033 ± 0.014 &
   &
  0.194 ± 0.008 &
   &
  \textbf{0.147} ± 0.010 &
   \\ \midrule
\textbf{KBIR} &
  0.251 ± 0.010 &
  \multirow{2}{*}{1.59\%} &
  0.047 ± 0.027 &
  \multirow{2}{*}{23.40\%} &
  0.189 ± 0.025 &
  \multirow{2}{*}{1.06\%} &
  0.133 ± 0.010 &
  \multirow{2}{*}{1.50\%} \\ \cmidrule(r){1-2} \cmidrule(lr){4-4} \cmidrule(lr){6-6} \cmidrule(lr){8-8}
\textbf{\begin{tabular}[c]{@{}c@{}}Graph Enhanced \\ KBIR\end{tabular}} &
  \textbf{0.255} ± 0.009 &
   &
  0.058 ± 0.016 &
   &
  0.191 ± 0.002 &
   &
  0.135 ± 0.007 &
   \\ \midrule
\textbf{Longformer} &
  0.235 ± 0.012 &
  \multirow{2}{*}{4.68\%} &
  0.043 ± 0.005 &
  \multirow{2}{*}{34.88\%} &
  0.198 ± 0.006 &
  \multirow{2}{*}{2.02\%} &
  0.141 ± 0.010 &
  \multirow{2}{*}{2.84\%} \\ \cmidrule(r){1-2} \cmidrule(lr){4-4} \cmidrule(lr){6-6} \cmidrule(lr){8-8}
\textbf{\begin{tabular}[c]{@{}c@{}}Graph Enhanced \\ Longformer\end{tabular}} &
  0.246 ± 0.014 &
   &
  0.068 ± 0.006 &
   &
  \textbf{0.202} ± 0.003 &
   &
  0.145 ± 0.012 &
   \\ \bottomrule
\end{tabular}%
}
\vspace{3mm}
\caption{Performance of models trained on SemEval-2010 train data and evaluated on the test splits of all the datasets.}
\label{tab:semeval-results}
\end{table}

The improvement of F1 scores observed in the LongFormer variant across all the experiments, highlights the usefulness of incorporating graph representation, even in cases where a PLM can represent longer sequences (4,096 vs 512). Our findings suggest that the training of PLMs with graph representation allows better understanding of long contexts and the development of better word representations by capturing both in context and out-of-context relationships between words, leading to significant improvements in the results across all the datasets.

\begin{table}[]
\centering
\resizebox{\textwidth}{!}{%
\begin{tabular}{@{}|c|c|c|c|c|c|c|c|c|@{}}
\toprule
\textbf{Model} &
  \textbf{SemEval-2010} &
  \textbf{\begin{tabular}[c]{@{}c@{}}\% \\ improve\end{tabular}} &
  \textbf{DUC-2001} &
  \textbf{\begin{tabular}[c]{@{}c@{}}\% \\ improve\end{tabular}} &
  \textbf{NUS} &
  \textbf{\begin{tabular}[c]{@{}c@{}}\% \\ improve\end{tabular}} &
  \textbf{LDKP3K} &
  \textbf{\begin{tabular}[c]{@{}c@{}}\% \\ improve\end{tabular}} \\ \midrule
\textbf{DistilBERT} &
  0.133 ± 0.008 &
  \multirow{2}{*}{9.02\%} &
  0.046 ± 0.003 &
  \multirow{2}{*}{19.57\%} &
  0.172 ± 0.004 &
  \multirow{2}{*}{13.37\%} &
  0.216 ± 0.005 &
  \multirow{2}{*}{5.56\%} \\ \cmidrule(r){1-2} \cmidrule(lr){4-4} \cmidrule(lr){6-6} \cmidrule(lr){8-8}
\textbf{\begin{tabular}[c]{@{}c@{}}Graph Enhanced \\ DistiBERT\end{tabular}} &
  0.145 ± 0.009 &
   &
  0.055 ± 0.004 &
   &
  0.195 ± 0.003 &
   &
  0.228 ± 0.007 &
   \\ \midrule
\textbf{BERT} &
  0.153 ± 0.008 &
  \multirow{2}{*}{9.80\%} &
  0.079 ± 0.008 &
  \multirow{2}{*}{56.96\%} &
  0.188 ± 0.003 &
  \multirow{2}{*}{19.68\%} &
  0.238 ± 0.003 &
  \multirow{2}{*}{5.88\%} \\ \cmidrule(r){1-2} \cmidrule(lr){4-4} \cmidrule(lr){6-6} \cmidrule(lr){8-8}
\textbf{\begin{tabular}[c]{@{}c@{}}Graph Enhanced \\ BERT\end{tabular}} &
  \textbf{0.168} ± 0.003 &
   &
  \textbf{0.124} ± 0.005 &
   &
  0.225 ± 0.007 &
   &
  0.252 ± 0.002 &
   \\ \midrule
\textbf{SciBERT} &
  0.146 ± 0.010 &
  \multirow{2}{*}{13.70\%} &
  0.017 ± 0.010 &
  \multirow{2}{*}{123.53\%} &
  0.190 ± 0.008 &
  \multirow{2}{*}{26.84\%} &
  0.245 ± 0.009 &
  \multirow{2}{*}{6.94\%} \\ \cmidrule(r){1-2} \cmidrule(lr){4-4} \cmidrule(lr){6-6} \cmidrule(lr){8-8}
\textbf{\begin{tabular}[c]{@{}c@{}}Graph Enhanced \\ SciBERT\end{tabular}} &
  0.166 ± 0.015 &
   &
  0.038 ± 0.015 &
   &
  \textbf{0.241} ± 0.006 &
   &
  0.262 ± 0.010 &
   \\ \midrule
\textbf{KBIR} &
  0.148 ± 0.008 &
  \multirow{2}{*}{8.11\%} &
  0.070 ± 0.010 &
  \multirow{2}{*}{17.14\%} &
  0.191 ± 0.011 &
  \multirow{2}{*}{9.95\%} &
  0.252 ± 0.009 &
  \multirow{2}{*}{5.16\%} \\ \cmidrule(r){1-2} \cmidrule(lr){4-4} \cmidrule(lr){6-6} \cmidrule(lr){8-8}
\textbf{\begin{tabular}[c]{@{}c@{}}Graph Enhanced \\ KBIR\end{tabular}} &
  0.160 ± 0.011 &
   &
  0.082 ± 0.011 &
   &
  0.210 ± 0.008 &
   &
  0.265 ± 0.011 &
   \\ \midrule
\textbf{Longformer} &
  0.146 ± 0.015 &
  \multirow{2}{*}{14.38\%} &
  0.085 ± 0.004 &
  \multirow{2}{*}{7.06\%} &
  0.207 ± 0.005 &
  \multirow{2}{*}{3.86\%} &
  0.260 ± 0.012 &
  \multirow{2}{*}{4.23\%} \\ \cmidrule(r){1-2} \cmidrule(lr){4-4} \cmidrule(lr){6-6} \cmidrule(lr){8-8}
\textbf{\begin{tabular}[c]{@{}c@{}}Graph Enhanced \\ Longformer\end{tabular}} &
  0.167 ± 0.011 &
   &
  0.091 ± 0.007 &
   &
  0.215 ± 0.004 &
   &
  \textbf{0.271} ± 0.008 &
   \\ \bottomrule
\end{tabular}%
}
\vspace{3mm}
\caption{Performance of models trained on LDKP3K train data and evaluated on the test splits of all the datasets.}
\label{tab:ldkp3k-results}
\end{table}


The results obtained by training the models on LDKP3K which has a larger training dataset is reported in Table \ref{tab:ldkp3k-results}. As expected, the availability of a larger training dataset results in a more significant improvement in performance. This is evident in the enhanced results across all datasets, including SemEval-2010, LDKP3K, and NUS, as reported in Table \ref{tab:ldkp3k-results}. These findings support our hypothesis that training the PLM and graph embeddings jointly results in a synergistic effect, which is further amplified by the availability of more data. Moreover, with the increase in the training data, the model can better capture the long-term dependencies present in the graph representation, leading to more effective learning.

In addition, when graph embeddings are integrated, the models demonstrate significant improvements in their performance on the news domain (DUC-2001), despite being fine-tuned on scientific documents. This suggests that the inclusion of graph embeddings enhances the models' ability to handle domain-specific distribution shifts. The domain-agnostic nature of graph embeddings makes them a promising approach for addressing challenges related to domain adaptation. Consequently, graph embeddings can serve as a valuable data augmentation technique, particularly in scenarios where domain-specific data is scarce.

Given that currently keyphrase extraction models heavily rely on training and evaluating models within the scientific domain, domain adaptation becomes a crucial aspect of interest. Our findings in this study strongly indicate that incorporating graph embeddings holds promise as a means to address the domain adaptation problem in keyphrase extraction. This highlights the potential for further exploration in utilizing graph embeddings for effectively tackling domain adaptation challenges in this field.


\subsection{Short Documents}



In this section, we present the results of our experiments on the Inspec dataset, which aim to assess whether incorporating graph embeddings can improve the performance of PLM-based models even for short documents that fit within the PLM's context. Table \ref{tab:inspec-results} summarizes the findings.

Our analysis of the results indicates that incorporating graph embeddings indeed enhances the PLM's representation even for short documents. This improvement is consistently observed across all evaluated models, including SciBERT, which was pre-trained on a domain-specific corpus, and KBIR, which underwent pretraining with task-specific objectives. Specifically, the addition of graph information leads to an average performance gain of more than 1 point in the F1 score. 

Based on our experimental observations, it is evident that incorporating graph embeddings enhances the representation of local context by capturing relationships that may not be readily apparent in sequential representations from PLMs. This is attributed to the unique portrayal of words in the graph, which enables the incorporation of connections that may not be fully captured by the PLM alone. The integration of graph and PLM representations synergistically enhances the embedding of local context, resulting in improved overall performance.

\begin{table}[]
\centering
\begin{tabular}{@{}|c|c|c|c|@{}}
\toprule
\textbf{Model} & \textbf{F1}   & \textbf{F1 (our approach)} & \textbf{\% improve} \\ \midrule
\textbf{DistilBERT}     & 0.506 ± 0.004 & 0.509 ± 0.005              & 0.59\%              \\ \midrule
\textbf{BERT}           & 0.514 ± 0.012 & 0.529 ± 0.013              & 2.92\%              \\ \midrule
\textbf{SciBERT}        & 0.518 ± 0.001 & 0.530 ± 0.006              & 2.32\%              \\ \midrule
\textbf{KBIR}           & 0.562 ± 0.012 & \textbf{0.570} ± 0.002              & 1.42\%              \\ \bottomrule
\end{tabular}
\vspace{3mm}
\caption{Performance of PLMs and Graph Enhance Sequence Tagger on Inspec.}
\label{tab:inspec-results}
\end{table}


\section{Case Studies}
In this section, we present concrete examples that demonstrate how our methodology enhances KPE from lengthy documents. To illustrate the effectiveness of our approach, we compare the performance of two models: KBIR, which is currently the state-of-the-art model in this task and has been fine-tuned on the Inspec dataset, and LongFormer, a high-performing model specifically designed for processing long documents, fine-tuned on SemEval-2010. We will contrast their results with those generated by one of our models, namely the Graph Enhanced LongFormer that has been trained on SemEval-2010.

\subsection{Case Study 1}
This example is taken from the SemEval-2010 test sample with ID `J-9'. The KPs that all methods were able to predict correctly are highlighted in red, while the KPs that only our model could predict correctly are highlighted in green. 

\begin{figure}[H]
    \centering
    \includegraphics[scale=0.6]{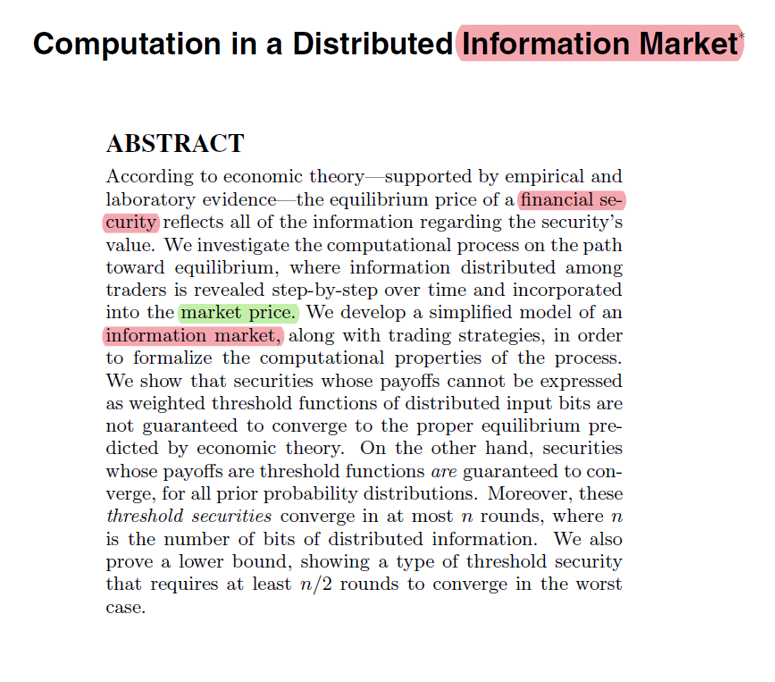}
    \caption{Case Study 1 - Title and Abstract}
    \label{fig:case-study-1-abstract}
\end{figure}



\begin{figure}[H]
    \centering
    \includegraphics[scale=0.33]{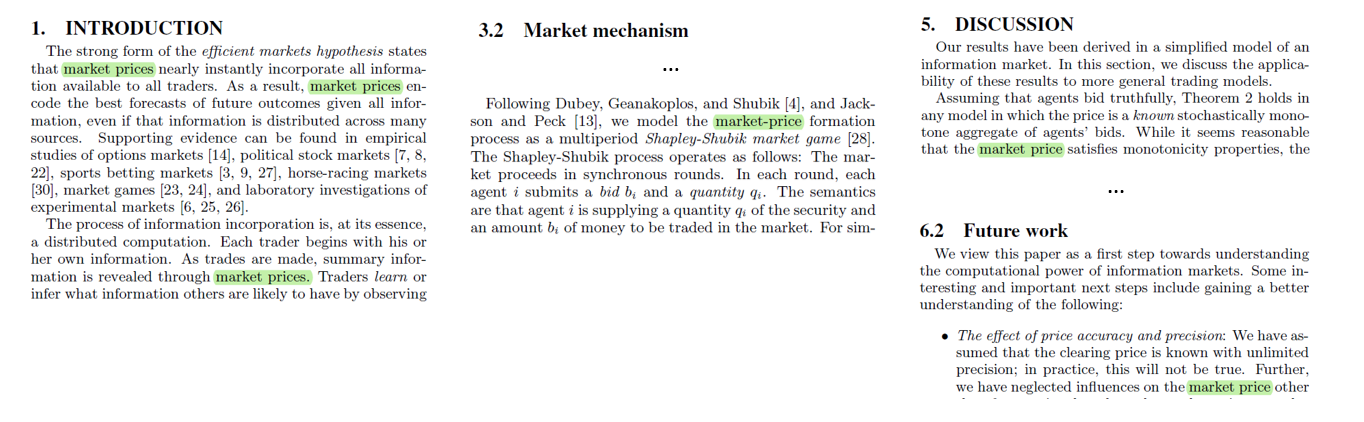}
    \caption{Case Study 1 - Excerpts from the document where the KPs are mentioned. The KPs that all methods were able to predict correctly are highlighted in red, while the KPs that only our model could predict correctly are highlighted in green.}
    \label{fig:case-study-1-rest}
\end{figure}

Figure \ref{fig:case-study-1-abstract} displays the title and abstract of the sample. As evident from the text, the phrases ``information market" and ``financial security" are repeated multiple times, including in the title, indicating their significance. However, the term ``market price" is mentioned only once, and its importance cannot be determined from its local context. Further analysis, as illustrated in Figure \ref{fig:case-study-1-rest}, reveals that although the term appears eight times in the document, it is not identified in the introduction, methodology, or discussion without the use of graph embeddings. 

This suggests that the KBIR model, which has a limited local context, cannot assess its importance. Even the longformer model, which can capture longer contexts of up to 4,096 tokens, fails to identify it as a keyphrase. In contrast, our graph-enhanced approach demonstrates a richer and more comprehensive understanding of the input contexts, providing the necessary information for the model to recognize ``market price" as a keyphrase.

In order for a human to identify the keyphrase ``market price", they would need to comprehend information from various sections of the paper, such as the abstract, introduction, related work, and methodology, and view the document as a cohesive whole. This holistic perspective is essential for comprehending lengthy documents, and it is precisely what graph embeddings aim to capture.

\subsection{Case Study 2}
This example is taken from the SemEval-2010 test sample with ID 'C-86'. The KPs that all methods were able to predict correctly are highlighted in red, while the KPs that only our model could predict correctly are highlighted in green. 

\begin{figure}[H]
    \centering
    \includegraphics[scale=0.42]{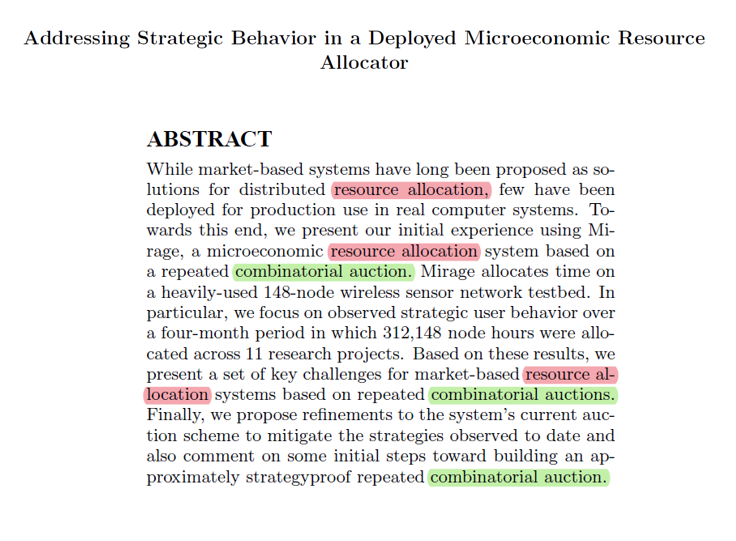}
    \caption{Case Study 2 - Title and Abstract.}
    \label{fig:case-study-2-abstract}
\end{figure}

\begin{figure}[H]
    \centering
    \includegraphics[scale=0.33]{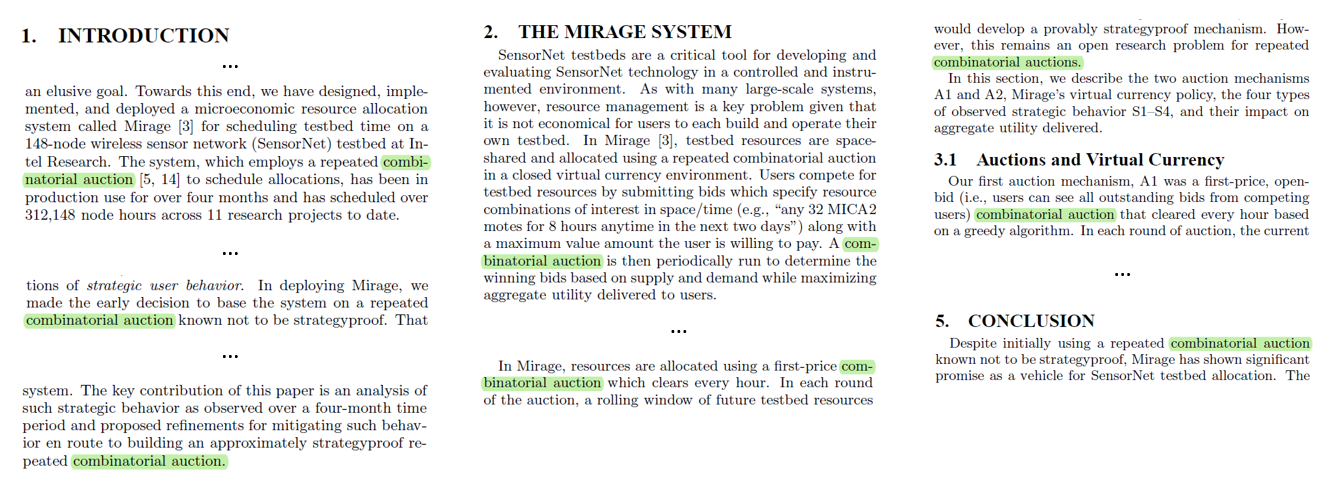}
    \caption{Case Study 2 - Excerpts from the document where the KPs are mentioned. The KPs that all methods were able to predict correctly are highlighted in red, while the KPs that only our model could predict correctly are highlighted in green.}
    \label{fig:case-study-2-rest}
\end{figure}

Figure \ref{fig:case-study-2-abstract} presents the title and abstract of the sample. In the given input context, ``resource allocation" is a keyphrase, while the importance of ``combinatorial auction" remains unclear to the KBIR and Longformer models, despite frequent references throughout the document. As shown in Figure \ref{fig:case-study-2-rest}, these models struggle to recognize the phrase, despite its appearance of 19 times in the paper, potentially due to their limited understanding of the global context.

In contrast, our graph-enhanced approach demonstrates superior performance by identifying ``combinatorial auction" as a keyphrase. This highlights the importance of a more comprehensive understanding of the document's global context, which is a critical advantage of our methodology.

\section{Limitations Conclusion and Future Work}
\label{conclusion}
\subsection{Limitations}



The integration of graph embeddings has proven to enhance the performance of Keyphrase Extraction (KPE) models for lengthy documents by providing a more comprehensive understanding of a word's context within the entire text. However, it should be acknowledged that this computational approach adds complexity to the pipeline. To assess the impact of this limitation, we measured the average time added to the KPE pipeline. We found that incorporating graph embeddings results in a time increase ranging from 20\% to 50\%, depending on the document's length.

Despite the additional computational time, we consider the improved performance achieved through our graph-enhanced approach as a valuable trade-off. By enabling a richer and more global understanding of the document's context, our methodology enhances the accuracy of KPE for lengthy texts, which is particularly important in various real-world applications.

It is important to acknowledge that our study focuses on pre-trained language models (PLMs) that fit within our computational budget. However, it is worth noting that significant progress has been made in the domain of Large Language Models (LLMs), which can process text sequences up to 32,768 tokens in length \cite{openai2023gpt}. Given the limitations of our study, we cannot assess the performance of our approach on such LLMs. Nonetheless, we anticipate that our graph-enhanced approach can be adapted and applied to these models to further enhance their performance in KPE for lengthy texts.

\subsection{Conclusion and Future Work}

In this study, we introduced the ``Graph Enhanced Sequence Tagger", which effectively combines the strengths of pre-trained language models (PLMs) and graph representations to enhance Keyphrase Extraction (KPE) in long documents. Our experimental results convincingly demonstrated that our proposed graph-enhanced approach outperforms state-of-the-art models, irrespective of document length. Notably, even specialized PLMs like Longformer, known for their proficiency in capturing long contexts in lengthy documents, benefited from our approach. The joint fine-tuning of the PLM and graph embedding yields a synergistic effect, enabling effective capture of long-term relationships and incorporation of topic-specific information. Furthermore, the promising results on out-of-domain datasets highlight the potential of our approach in domain adaptation.

In our future work, we aim to explore additional graph representations, such as lexical and/or syntactic dependency graphs, to further enhance the performance of our method. We also intend to investigate the utilization of self-supervised training objectives that are more closely aligned with the KPE task, aiming to enrich the learned graph representation. Lastly, we plan to explore the problem of domain adaptation for keyphrase extraction and generation by incorporating graph representations into relevant approaches.

\bibliography{sn-bibliography}


\begin{thebibliography}{62}
\ifx \bisbn   \undefined \def \bisbn  #1{ISBN #1}\fi
\ifx \binits  \undefined \def \binits#1{#1}\fi
\ifx \bauthor  \undefined \def \bauthor#1{#1}\fi
\ifx \batitle  \undefined \def \batitle#1{#1}\fi
\ifx \bjtitle  \undefined \def \bjtitle#1{#1}\fi
\ifx \bvolume  \undefined \def \bvolume#1{\textbf{#1}}\fi
\ifx \byear  \undefined \def \byear#1{#1}\fi
\ifx \bissue  \undefined \def \bissue#1{#1}\fi
\ifx \bfpage  \undefined \def \bfpage#1{#1}\fi
\ifx \blpage  \undefined \def \blpage #1{#1}\fi
\ifx \burl  \undefined \def \burl#1{\textsf{#1}}\fi
\ifx \doiurl  \undefined \def \doiurl#1{\url{https://doi.org/#1}}\fi
\ifx \betal  \undefined \def \betal{\textit{et al.}}\fi
\ifx \binstitute  \undefined \def \binstitute#1{#1}\fi
\ifx \binstitutionaled  \undefined \def \binstitutionaled#1{#1}\fi
\ifx \bctitle  \undefined \def \bctitle#1{#1}\fi
\ifx \beditor  \undefined \def \beditor#1{#1}\fi
\ifx \bpublisher  \undefined \def \bpublisher#1{#1}\fi
\ifx \bbtitle  \undefined \def \bbtitle#1{#1}\fi
\ifx \bedition  \undefined \def \bedition#1{#1}\fi
\ifx \bseriesno  \undefined \def \bseriesno#1{#1}\fi
\ifx \blocation  \undefined \def \blocation#1{#1}\fi
\ifx \bsertitle  \undefined \def \bsertitle#1{#1}\fi
\ifx \bsnm \undefined \def \bsnm#1{#1}\fi
\ifx \bsuffix \undefined \def \bsuffix#1{#1}\fi
\ifx \bparticle \undefined \def \bparticle#1{#1}\fi
\ifx \barticle \undefined \def \barticle#1{#1}\fi
\bibcommenthead
\ifx \bconfdate \undefined \def \bconfdate #1{#1}\fi
\ifx \botherref \undefined \def \botherref #1{#1}\fi
\ifx \url \undefined \def \url#1{\textsf{#1}}\fi
\ifx \bchapter \undefined \def \bchapter#1{#1}\fi
\ifx \bbook \undefined \def \bbook#1{#1}\fi
\ifx \bcomment \undefined \def \bcomment#1{#1}\fi
\ifx \oauthor \undefined \def \oauthor#1{#1}\fi
\ifx \citeauthoryear \undefined \def \citeauthoryear#1{#1}\fi
\ifx \endbibitem  \undefined \def \endbibitem {}\fi
\ifx \bconflocation  \undefined \def \bconflocation#1{#1}\fi
\ifx \arxivurl  \undefined \def \arxivurl#1{\textsf{#1}}\fi
\csname PreBibitemsHook\endcsname

\bibitem{hulth2006study}
\begin{bchapter}
\bauthor{\bsnm{Hulth}, \binits{A.}},
\bauthor{\bsnm{Megyesi}, \binits{B.}}:
\bctitle{A study on automatically extracted keywords in text categorization}.
In: \bbtitle{Proceedings of the 21st International Conference on Computational
  Linguistics and 44th Annual Meeting of the Association for Computational
  Linguistics},
pp. \bfpage{537}--\blpage{544}
(\byear{2006})
\end{bchapter}
\endbibitem

\bibitem{hammouda2005corephrase}
\begin{bchapter}
\bauthor{\bsnm{Hammouda}, \binits{K.M.}},
\bauthor{\bsnm{Matute}, \binits{D.N.}},
\bauthor{\bsnm{Kamel}, \binits{M.S.}}:
\bctitle{Corephrase: Keyphrase extraction for document clustering}.
In: \bbtitle{International Workshop on Machine Learning and Data Mining in
  Pattern Recognition},
pp. \bfpage{265}--\blpage{274}
(\byear{2005}).
\bcomment{Springer}
\end{bchapter}
\endbibitem

\bibitem{summarization-example-1}
\begin{bchapter}
\bauthor{\bsnm{Qazvinian}, \binits{V.}},
\bauthor{\bsnm{Radev}, \binits{D.}},
\bauthor{\bsnm{{\"O}zg{\"u}r}, \binits{A.}}:
\bctitle{Citation summarization through keyphrase extraction}.
In: \bbtitle{Proceedings of the 23rd International Conference on Computational
  Linguistics (COLING 2010)},
pp. \bfpage{895}--\blpage{903}
(\byear{2010})
\end{bchapter}
\endbibitem

\bibitem{summarization-example-2}
\begin{barticle}
\bauthor{\bsnm{Zhang}, \binits{Y.}},
\bauthor{\bsnm{Zincir-Heywood}, \binits{N.}},
\bauthor{\bsnm{Milios}, \binits{E.}}:
\batitle{World wide web site summarization}.
\bjtitle{Web Intelligence and Agent Systems: An International Journal}
\bvolume{2}(\bissue{1}),
\bfpage{39}--\blpage{53}
(\byear{2004})
\end{barticle}
\endbibitem

\bibitem{ie-example}
\begin{barticle}
\bauthor{\bsnm{Gutwin}, \binits{C.}},
\bauthor{\bsnm{Paynter}, \binits{G.}},
\bauthor{\bsnm{Witten}, \binits{I.}},
\bauthor{\bsnm{Nevill-Manning}, \binits{C.}},
\bauthor{\bsnm{Frank}, \binits{E.}}:
\batitle{Improving browsing in digital libraries with keyphrase indexes}.
\bjtitle{Decision Support Systems}
\bvolume{27}(\bissue{1-2}),
\bfpage{81}--\blpage{104}
(\byear{1999})
\end{barticle}
\endbibitem

\bibitem{query-expansion-example}
\begin{bchapter}
\bauthor{\bsnm{Song}, \binits{I.Y.}},
\bauthor{\bsnm{Allen}, \binits{R.B.}},
\bauthor{\bsnm{Obradovic}, \binits{Z.}},
\bauthor{\bsnm{Song}, \binits{M.}}:
\bctitle{Keyphrase extraction-based query expansion in digital libraries}.
In: \bbtitle{Proceedings of the 6th ACM/IEEE-CS Joint Conference on Digital
  Libraries (JCDL'06)},
pp. \bfpage{202}--\blpage{209}
(\byear{2006}).
\bcomment{IEEE}
\end{bchapter}
\endbibitem

\bibitem{jones1999phrasier}
\begin{bchapter}
\bauthor{\bsnm{Jones}, \binits{S.}},
\bauthor{\bsnm{Staveley}, \binits{M.S.}}:
\bctitle{Phrasier: a system for interactive document retrieval using
  keyphrases}.
In: \bbtitle{Proceedings of the 22nd Annual International ACM SIGIR Conference
  on Research and Development in Information Retrieval},
pp. \bfpage{160}--\blpage{167}
(\byear{1999})
\end{bchapter}
\endbibitem

\bibitem{keyphrasification}
\begin{bchapter}
\bauthor{\bsnm{Meng}, \binits{R.}},
\bauthor{\bsnm{Mahata}, \binits{D.}},
\bauthor{\bsnm{Boudin}, \binits{F.}}:
\bctitle{From fundamentals to recent advances: A tutorial on
  keyphrasification}.
In: \bbtitle{Advances in Information Retrieval: 44th European Conference on IR
  Research, ECIR 2022, Stavanger, Norway, April 10--14, 2022, Proceedings, Part
  II},
pp. \bfpage{582}--\blpage{588}
(\byear{2022}).
\bcomment{Springer}
\end{bchapter}
\endbibitem

\bibitem{ccano2019keyphrasesurvey}
\begin{bchapter}
\bauthor{\bsnm{{\c{C}}ano}, \binits{E.}},
\bauthor{\bsnm{Bojar}, \binits{O.}}:
\bctitle{Keyphrase generation: A multi-aspect survey}.
In: \bbtitle{2019 25th Conference of Open Innovations Association (FRUCT)},
pp. \bfpage{85}--\blpage{94}
(\byear{2019}).
\bcomment{IEEE}
\end{bchapter}
\endbibitem

\bibitem{kesurvey2014}
\begin{bchapter}
\bauthor{\bsnm{Hasan}, \binits{K.S.}},
\bauthor{\bsnm{Ng}, \binits{V.}}:
\bctitle{Automatic keyphrase extraction: A survey of the state of the art}.
In: \bbtitle{Proceedings of the 52nd Annual Meeting of the Association for
  Computational Linguistics (Volume 1: Long Papers)},
pp. \bfpage{1262}--\blpage{1273}
(\byear{2014})
\end{bchapter}
\endbibitem

\bibitem{graph-based-ke}
\begin{bchapter}
\bauthor{\bsnm{Mothe}, \binits{J.}},
\bauthor{\bsnm{Ramiandrisoa}, \binits{F.}},
\bauthor{\bsnm{Rasolomanana}, \binits{M.}}:
\bctitle{Automatic keyphrase extraction using graph-based methods}.
In: \bbtitle{Proceedings of the 33rd Annual ACM Symposium on Applied
  Computing},
pp. \bfpage{728}--\blpage{730}
(\byear{2018})
\end{bchapter}
\endbibitem

\bibitem{boudin2013comparison}
\begin{bchapter}
\bauthor{\bsnm{Boudin}, \binits{F.}}:
\bctitle{A comparison of centrality measures for graph-based keyphrase
  extraction}.
In: \bbtitle{Proceedings of the Sixth International Joint Conference on Natural
  Language Processing},
pp. \bfpage{834}--\blpage{838}
(\byear{2013})
\end{bchapter}
\endbibitem

\bibitem{hasan2014automatic}
\begin{bchapter}
\bauthor{\bsnm{Hasan}, \binits{K.S.}},
\bauthor{\bsnm{Ng}, \binits{V.}}:
\bctitle{Automatic keyphrase extraction: A survey of the state of the art}.
In: \bbtitle{Proceedings of the 52nd Annual Meeting of the Association for
  Computational Linguistics (Volume 1: Long Papers)},
pp. \bfpage{1262}--\blpage{1273}
(\byear{2014})
\end{bchapter}
\endbibitem

\bibitem{patel2019exploring}
\begin{bchapter}
\bauthor{\bsnm{Patel}, \binits{K.}},
\bauthor{\bsnm{Caragea}, \binits{C.}}:
\bctitle{Exploring word embeddings in crf-based keyphrase extraction from
  research papers}.
In: \bbtitle{Proceedings of the 10th International Conference on Knowledge
  Capture},
pp. \bfpage{37}--\blpage{44}
(\byear{2019})
\end{bchapter}
\endbibitem

\bibitem{word2vec}
\begin{botherref}
\oauthor{\bsnm{Mikolov}, \binits{T.}},
\oauthor{\bsnm{Chen}, \binits{K.}},
\oauthor{\bsnm{Corrado}, \binits{G.}},
\oauthor{\bsnm{Dean}, \binits{J.}}:
Efficient Estimation of Word Representations in Vector Space
(2013)
\end{botherref}
\endbibitem

\bibitem{glove}
\begin{bchapter}
\bauthor{\bsnm{Pennington}, \binits{J.}},
\bauthor{\bsnm{Socher}, \binits{R.}},
\bauthor{\bsnm{Manning}, \binits{C.}}:
\bctitle{{G}lo{V}e: Global vectors for word representation}.
In: \bbtitle{Proceedings of the 2014 Conference on Empirical Methods in Natural
  Language Processing ({EMNLP})},
pp. \bfpage{1532}--\blpage{1543}.
\bpublisher{Association for Computational Linguistics},
\blocation{Doha, Qatar}
(\byear{2014}).
\doiurl{10.3115/v1/D14-1162}.
\burl{https://aclanthology.org/D14-1162}
\end{bchapter}
\endbibitem

\bibitem{sahrawat2020keyphrase}
\begin{bchapter}
\bauthor{\bsnm{Sahrawat}, \binits{D.}},
\bauthor{\bsnm{Mahata}, \binits{D.}},
\bauthor{\bsnm{Zhang}, \binits{H.}},
\bauthor{\bsnm{Kulkarni}, \binits{M.}},
\bauthor{\bsnm{Sharma}, \binits{A.}},
\bauthor{\bsnm{Gosangi}, \binits{R.}},
\bauthor{\bsnm{Stent}, \binits{A.}},
\bauthor{\bsnm{Kumar}, \binits{Y.}},
\bauthor{\bsnm{Shah}, \binits{R.R.}},
\bauthor{\bsnm{Zimmermann}, \binits{R.}}:
\bctitle{Keyphrase extraction as sequence labeling using contextualized
  embeddings}.
In: \bbtitle{European Conference on Information Retrieval},
pp. \bfpage{328}--\blpage{335}
(\byear{2020}).
\bcomment{Springer}
\end{bchapter}
\endbibitem

\bibitem{bert}
\begin{botherref}
\oauthor{\bsnm{Devlin}, \binits{J.}},
\oauthor{\bsnm{Chang}, \binits{M.-W.}},
\oauthor{\bsnm{Lee}, \binits{K.}},
\oauthor{\bsnm{Toutanova}, \binits{K.}}:
BERT: Pre-training of Deep Bidirectional Transformers for Language
  Understanding.
arXiv
(2018).
\doiurl{10.48550/ARXIV.1810.04805}.
\url{https://arxiv.org/abs/1810.04805}
\end{botherref}
\endbibitem

\bibitem{bilstm}
\begin{botherref}
\oauthor{\bsnm{Huang}, \binits{Z.}},
\oauthor{\bsnm{Xu}, \binits{W.}},
\oauthor{\bsnm{Yu}, \binits{K.}}:
Bidirectional LSTM-CRF Models for Sequence Tagging.
arXiv
(2015).
\doiurl{10.48550/ARXIV.1508.01991}.
\url{https://arxiv.org/abs/1508.01991}
\end{botherref}
\endbibitem

\bibitem{kbir}
\begin{bchapter}
\bauthor{\bsnm{Kulkarni}, \binits{M.}},
\bauthor{\bsnm{Mahata}, \binits{D.}},
\bauthor{\bsnm{Arora}, \binits{R.}},
\bauthor{\bsnm{Bhowmik}, \binits{R.}}:
\bctitle{Learning rich representation of keyphrases from text}.
In: \bbtitle{Findings of the Association for Computational Linguistics: NAACL
  2022},
pp. \bfpage{891}--\blpage{906}.
\bpublisher{Association for Computational Linguistics},
\blocation{Seattle, United States}
(\byear{2022}).
\doiurl{10.18653/v1/2022.findings-naacl.67}.
\burl{https://aclanthology.org/2022.findings-naacl.67}
\end{bchapter}
\endbibitem

\bibitem{park2020scientific}
\begin{bchapter}
\bauthor{\bsnm{Park}, \binits{S.}},
\bauthor{\bsnm{Caragea}, \binits{C.}}:
\bctitle{Scientific keyphrase identification and classification by pre-trained
  language models intermediate task transfer learning}.
In: \bbtitle{Proceedings of the 28th International Conference on Computational
  Linguistics},
pp. \bfpage{5409}--\blpage{5419}
(\byear{2020})
\end{bchapter}
\endbibitem

\bibitem{gcn}
\begin{botherref}
\oauthor{\bsnm{Kipf}, \binits{T.N.}},
\oauthor{\bsnm{Welling}, \binits{M.}}:
Semi-Supervised Classification with Graph Convolutional Networks.
arXiv
(2016).
\doiurl{10.48550/ARXIV.1609.02907}.
\url{https://arxiv.org/abs/1609.02907}
\end{botherref}
\endbibitem

\bibitem{sasake}
\begin{bchapter}
\bauthor{\bsnm{Santosh}, \binits{T.y.s.s.}},
\bauthor{\bsnm{Kumar~Sanyal}, \binits{D.}},
\bauthor{\bsnm{Bhowmick}, \binits{P.K.}},
\bauthor{\bsnm{Das}, \binits{P.P.}}:
\bctitle{{S}a{SAKE}: Syntax and semantics aware keyphrase extraction from
  research papers}.
In: \bbtitle{Proceedings of the 28th International Conference on Computational
  Linguistics},
pp. \bfpage{5372}--\blpage{5383}.
\bpublisher{International Committee on Computational Linguistics},
\blocation{Barcelona, Spain (Online)}
(\byear{2020}).
\doiurl{10.18653/v1/2020.coling-main.469}.
\burl{https://aclanthology.org/2020.coling-main.469}
\end{bchapter}
\endbibitem

\bibitem{phraseformer}
\begin{botherref}
\oauthor{\bsnm{Nikzad-Khasmakhi}, \binits{N.}},
\oauthor{\bsnm{Feizi-Derakhshi}, \binits{M.-R.}},
\oauthor{\bsnm{Asgari-Chenaghlu}, \binits{M.}},
\oauthor{\bsnm{Balafar}, \binits{M.-A.}},
\oauthor{\bsnm{Feizi-Derakhshi}, \binits{A.-R.}},
\oauthor{\bsnm{Rahkar-Farshi}, \binits{T.}},
\oauthor{\bsnm{Ramezani}, \binits{M.}},
\oauthor{\bsnm{Jahanbakhsh-Nagadeh}, \binits{Z.}},
\oauthor{\bsnm{Zafarani-Moattar}, \binits{E.}},
\oauthor{\bsnm{Ranjbar-Khadivi}, \binits{M.}}:
Phraseformer: Multimodal Key-phrase Extraction using Transformer and Graph
  Embedding.
arXiv
(2021).
\doiurl{10.48550/ARXIV.2106.04939}.
\url{https://arxiv.org/abs/2106.04939}
\end{botherref}
\endbibitem

\bibitem{pagerank}
\begin{bchapter}
\bauthor{\bsnm{Page}, \binits{L.}},
\bauthor{\bsnm{Brin}, \binits{S.}},
\bauthor{\bsnm{Motwani}, \binits{R.}},
\bauthor{\bsnm{Winograd}, \binits{T.}}:
\bctitle{The pagerank citation ranking : Bringing order to the web}.
In: \bbtitle{The Web Conference}
(\byear{1999})
\end{bchapter}
\endbibitem

\bibitem{textrank}
\begin{bchapter}
\bauthor{\bsnm{Mihalcea}, \binits{R.}},
\bauthor{\bsnm{Tarau}, \binits{P.}}:
\bctitle{{T}ext{R}ank: Bringing order into text}.
In: \bbtitle{Proceedings of the 2004 Conference on Empirical Methods in Natural
  Language Processing},
pp. \bfpage{404}--\blpage{411}.
\bpublisher{Association for Computational Linguistics},
\blocation{Barcelona, Spain}
(\byear{2004}).
\burl{https://aclanthology.org/W04-3252}
\end{bchapter}
\endbibitem

\bibitem{topicrank}
\begin{bchapter}
\bauthor{\bsnm{Bougouin}, \binits{A.}},
\bauthor{\bsnm{Boudin}, \binits{F.}},
\bauthor{\bsnm{Daille}, \binits{B.}}:
\bctitle{{T}opic{R}ank: Graph-based topic ranking for keyphrase extraction}.
In: \bbtitle{Proceedings of the Sixth International Joint Conference on Natural
  Language Processing},
pp. \bfpage{543}--\blpage{551}.
\bpublisher{Asian Federation of Natural Language Processing},
\blocation{Nagoya, Japan}
(\byear{2013}).
\burl{https://aclanthology.org/I13-1062}
\end{bchapter}
\endbibitem

\bibitem{wang2014corpus}
\begin{bchapter}
\bauthor{\bsnm{Wang}, \binits{R.}},
\bauthor{\bsnm{Liu}, \binits{W.}},
\bauthor{\bsnm{McDonald}, \binits{C.}}:
\bctitle{Corpus-independent generic keyphrase extraction using word embedding
  vectors}.
In: \bbtitle{Software Engineering Research Conference},
vol. \bseriesno{39},
pp. \bfpage{1}--\blpage{8}
(\byear{2014})
\end{bchapter}
\endbibitem

\bibitem{mahata2018key2vec}
\begin{bchapter}
\bauthor{\bsnm{Mahata}, \binits{D.}},
\bauthor{\bsnm{Kuriakose}, \binits{J.}},
\bauthor{\bsnm{Shah}, \binits{R.}},
\bauthor{\bsnm{Zimmermann}, \binits{R.}}:
\bctitle{Key2vec: Automatic ranked keyphrase extraction from scientific
  articles using phrase embeddings}.
In: \bbtitle{Proceedings of the 2018 Conference of the North American Chapter
  of the Association for Computational Linguistics: Human Language
  Technologies, Volume 2 (Short Papers)},
pp. \bfpage{634}--\blpage{639}
(\byear{2018})
\end{bchapter}
\endbibitem

\bibitem{mahata2018theme}
\begin{bchapter}
\bauthor{\bsnm{Mahata}, \binits{D.}},
\bauthor{\bsnm{Shah}, \binits{R.R.}},
\bauthor{\bsnm{Kuriakose}, \binits{J.}},
\bauthor{\bsnm{Zimmermann}, \binits{R.}},
\bauthor{\bsnm{Talburt}, \binits{J.R.}}:
\bctitle{Theme-weighted ranking of keywords from text documents using phrase
  embeddings}.
In: \bbtitle{2018 IEEE Conference on Multimedia Information Processing and
  Retrieval (MIPR)},
pp. \bfpage{184}--\blpage{189}
(\byear{2018}).
\bcomment{IEEE}
\end{bchapter}
\endbibitem

\bibitem{bennani2018simple}
\begin{botherref}
\oauthor{\bsnm{Bennani-Smires}, \binits{K.}},
\oauthor{\bsnm{Musat}, \binits{C.}},
\oauthor{\bsnm{Hossmann}, \binits{A.}},
\oauthor{\bsnm{Baeriswyl}, \binits{M.}},
\oauthor{\bsnm{Jaggi}, \binits{M.}}:
Simple unsupervised keyphrase extraction using sentence embeddings.
arXiv preprint arXiv:1801.04470
(2018)
\end{botherref}
\endbibitem

\bibitem{hulth2003}
\begin{bchapter}
\bauthor{\bsnm{Hulth}, \binits{A.}}:
\bctitle{Improved automatic keyword extraction given more linguistic
  knowledge}.
In: \bbtitle{Proceedings of the 2003 Conference on Empirical Methods in Natural
  Language Processing}.
\bsertitle{EMNLP '03},
pp. \bfpage{216}--\blpage{223}.
\bpublisher{Association for Computational Linguistics},
\blocation{USA}
(\byear{2003}).
\doiurl{10.3115/1119355.1119383}.
\burl{https://doi.org/10.3115/1119355.1119383}
\end{bchapter}
\endbibitem

\bibitem{kim-kan-2009-examining}
\begin{bchapter}
\bauthor{\bsnm{Kim}, \binits{S.N.}},
\bauthor{\bsnm{Kan}, \binits{M.-Y.}}:
\bctitle{Re-examining automatic keyphrase extraction approaches in scientific
  articles}.
In: \bbtitle{Proceedings of the Workshop on Multiword Expressions:
  Identification, Interpretation, Disambiguation and Applications ({MWE}
  2009)},
pp. \bfpage{9}--\blpage{16}.
\bpublisher{Association for Computational Linguistics},
\blocation{Singapore}
(\byear{2009}).
\burl{https://aclanthology.org/W09-2902}
\end{bchapter}
\endbibitem

\bibitem{nguyen-kan-2007}
\begin{bchapter}
\bauthor{\bsnm{Nguyen}, \binits{T.D.}},
\bauthor{\bsnm{Kan}, \binits{M.-Y.}}:
\bctitle{Keyphrase extraction in scientific publications}.
In: \beditor{\bsnm{Goh}, \binits{D.H.-L.}},
\beditor{\bsnm{Cao}, \binits{T.H.}},
\beditor{\bsnm{S{\o}lvberg}, \binits{I.T.}},
\beditor{\bsnm{Rasmussen}, \binits{E.}} (eds.)
\bbtitle{Asian Digital Libraries. Looking Back 10 Years and Forging New
  Frontiers},
pp. \bfpage{317}--\blpage{326}.
\bpublisher{Springer},
\blocation{Berlin, Heidelberg}
(\byear{2007})
\end{bchapter}
\endbibitem

\bibitem{Gollapalli-Li-Yang-2017}
\begin{botherref}
\oauthor{\bsnm{Gollapalli}, \binits{S.D.}},
\oauthor{\bsnm{Li}, \binits{X.-l.}},
\oauthor{\bsnm{Yang}, \binits{P.}}:
Incorporating expert knowledge into keyphrase extraction.
Proceedings of the AAAI Conference on Artificial Intelligence
\textbf{31}(1)
(2017).
\doiurl{10.1609/aaai.v31i1.10986}
\end{botherref}
\endbibitem

\bibitem{alzaidy-2019}
\begin{bchapter}
\bauthor{\bsnm{Alzaidy}, \binits{R.}},
\bauthor{\bsnm{Caragea}, \binits{C.}},
\bauthor{\bsnm{Giles}, \binits{C.L.}}:
\bctitle{Bi-lstm-crf sequence labeling for keyphrase extraction from scholarly
  documents}.
In: \bbtitle{The World Wide Web Conference}.
\bsertitle{WWW '19},
pp. \bfpage{2551}--\blpage{2557}.
\bpublisher{Association for Computing Machinery},
\blocation{New York, NY, USA}
(\byear{2019}).
\doiurl{10.1145/3308558.3313642}.
\burl{https://doi.org/10.1145/3308558.3313642}
\end{bchapter}
\endbibitem

\bibitem{attention-is-all-you-need}
\begin{botherref}
\oauthor{\bsnm{Vaswani}, \binits{A.}},
\oauthor{\bsnm{Shazeer}, \binits{N.}},
\oauthor{\bsnm{Parmar}, \binits{N.}},
\oauthor{\bsnm{Uszkoreit}, \binits{J.}},
\oauthor{\bsnm{Jones}, \binits{L.}},
\oauthor{\bsnm{Gomez}, \binits{A.N.}},
\oauthor{\bsnm{Kaiser}, \binits{L.}},
\oauthor{\bsnm{Polosukhin}, \binits{I.}}:
Attention Is All You Need.
arXiv
(2017).
\doiurl{10.48550/ARXIV.1706.03762}.
\url{https://arxiv.org/abs/1706.03762}
\end{botherref}
\endbibitem

\bibitem{transkp}
\begin{botherref}
\oauthor{\bsnm{Rungta}, \binits{M.}},
\oauthor{\bsnm{Kumar}, \binits{R.}},
\oauthor{\bsnm{Dhaliwal}, \binits{M.P.}},
\oauthor{\bsnm{Tiwari}, \binits{H.}},
\oauthor{\bsnm{Vala}, \binits{V.}}:
Transkp: Transformer based key-phrase extraction.
2020 International Joint Conference on Neural Networks (IJCNN),
1--7
(2020)
\end{botherref}
\endbibitem

\bibitem{tnt-kid}
\begin{barticle}
\bauthor{\bsnm{Martinc}, \binits{M.}},
\bauthor{\bsnm{{\v{S}}krlj}, \binits{B.}},
\bauthor{\bsnm{Pollak}, \binits{S.}}:
\batitle{{TNT}-{KID}: Transformer-based neural tagger for keyword
  identification}.
\bjtitle{Natural Language Engineering}
\bvolume{28}(\bissue{4}),
\bfpage{409}--\blpage{448}
(\byear{2021}).
\doiurl{10.1017/s1351324921000127}
\end{barticle}
\endbibitem

\bibitem{divgraphpointer}
\begin{bchapter}
\bauthor{\bsnm{Sun}, \binits{Z.}},
\bauthor{\bsnm{Tang}, \binits{J.}},
\bauthor{\bsnm{Du}, \binits{P.}},
\bauthor{\bsnm{Deng}, \binits{Z.-H.}},
\bauthor{\bsnm{Nie}, \binits{J.-Y.}}:
\bctitle{{DivGraphPointer}}.
In: \bbtitle{Proceedings of the 42nd International {ACM} {SIGIR} Conference on
  Research and Development in Information Retrieval}.
\bpublisher{{ACM}}, \blocation{???}
(\byear{2019}).
\doiurl{10.1145/3331184.3331219}.
\burl{https://doi.org/10.1145\%2F3331184.3331219}
\end{bchapter}
\endbibitem

\bibitem{heter-graph-kpg}
\begin{botherref}
\oauthor{\bsnm{Ye}, \binits{J.}},
\oauthor{\bsnm{Cai}, \binits{R.}},
\oauthor{\bsnm{Gui}, \binits{T.}},
\oauthor{\bsnm{Zhang}, \binits{Q.}}:
Heterogeneous Graph Neural Networks for Keyphrase Generation.
arXiv
(2021).
\doiurl{10.48550/ARXIV.2109.04703}.
\url{https://arxiv.org/abs/2109.04703}
\end{botherref}
\endbibitem

\bibitem{deepwalk}
\begin{bchapter}
\bauthor{\bsnm{Perozzi}, \binits{B.}},
\bauthor{\bsnm{Al-Rfou}, \binits{R.}},
\bauthor{\bsnm{Skiena}, \binits{S.}}:
\bctitle{{DeepWalk}}.
In: \bbtitle{Proceedings of the 20th {ACM} {SIGKDD} International Conference on
  Knowledge Discovery and Data Mining}.
\bpublisher{{ACM}}, \blocation{???}
(\byear{2014}).
\doiurl{10.1145/2623330.2623732}.
\burl{https://arxiv.org/abs/1403.6652}
\end{bchapter}
\endbibitem

\bibitem{node2vec}
\begin{botherref}
\oauthor{\bsnm{Grover}, \binits{A.}},
\oauthor{\bsnm{Leskovec}, \binits{J.}}:
node2vec: Scalable Feature Learning for Networks
(2016)
\end{botherref}
\endbibitem

\bibitem{sagegraph}
\begin{botherref}
\oauthor{\bsnm{Hamilton}, \binits{W.L.}},
\oauthor{\bsnm{Ying}, \binits{R.}},
\oauthor{\bsnm{Leskovec}, \binits{J.}}:
Inductive Representation Learning on Large Graphs.
arXiv
(2017).
\doiurl{10.48550/ARXIV.1706.02216}.
\url{https://arxiv.org/abs/1706.02216}
\end{botherref}
\endbibitem

\bibitem{gat}
\begin{botherref}
\oauthor{\bsnm{Veličković}, \binits{P.}},
\oauthor{\bsnm{Cucurull}, \binits{G.}},
\oauthor{\bsnm{Casanova}, \binits{A.}},
\oauthor{\bsnm{Romero}, \binits{A.}},
\oauthor{\bsnm{Liò}, \binits{P.}},
\oauthor{\bsnm{Bengio}, \binits{Y.}}:
Graph Attention Networks.
arXiv
(2017).
\doiurl{10.48550/ARXIV.1710.10903}.
\url{https://arxiv.org/abs/1710.10903}
\end{botherref}
\endbibitem

\bibitem{zhang2022link}
\begin{botherref}
\oauthor{\bsnm{Zhang}, \binits{D.}},
\oauthor{\bsnm{Yin}, \binits{J.}},
\oauthor{\bsnm{Yu}, \binits{P.S.}}:
Link Prediction with Contextualized Self-Supervision
(2022)
\end{botherref}
\endbibitem

\bibitem{saxena2021nodesim}
\begin{botherref}
\oauthor{\bsnm{Saxena}, \binits{A.}},
\oauthor{\bsnm{Fletcher}, \binits{G.}},
\oauthor{\bsnm{Pechenizkiy}, \binits{M.}}:
NodeSim: Node Similarity based Network Embedding for Diverse Link Prediction
(2021)
\end{botherref}
\endbibitem

\bibitem{bigbird}
\begin{botherref}
\oauthor{\bsnm{Zaheer}, \binits{M.}},
\oauthor{\bsnm{Guruganesh}, \binits{G.}},
\oauthor{\bsnm{Dubey}, \binits{A.}},
\oauthor{\bsnm{Ainslie}, \binits{J.}},
\oauthor{\bsnm{Alberti}, \binits{C.}},
\oauthor{\bsnm{Ontanon}, \binits{S.}},
\oauthor{\bsnm{Pham}, \binits{P.}},
\oauthor{\bsnm{Ravula}, \binits{A.}},
\oauthor{\bsnm{Wang}, \binits{Q.}},
\oauthor{\bsnm{Yang}, \binits{L.}},
\oauthor{\bsnm{Ahmed}, \binits{A.}}:
Big bird: Transformers for longer sequences
(2020).
\doiurl{10.48550/ARXIV.2007.14062}
\end{botherref}
\endbibitem

\bibitem{longformer}
\begin{botherref}
\oauthor{\bsnm{Beltagy}, \binits{I.}},
\oauthor{\bsnm{Peters}, \binits{M.E.}},
\oauthor{\bsnm{Cohan}, \binits{A.}}:
Longformer: The long-document transformer.
arXiv:2004.05150
(2020)
\end{botherref}
\endbibitem

\bibitem{classification-of-long-documents}
\begin{botherref}
\oauthor{\bsnm{Park}, \binits{H.H.}},
\oauthor{\bsnm{Vyas}, \binits{Y.}},
\oauthor{\bsnm{Shah}, \binits{K.}}:
Efficient Classification of Long Documents Using Transformers.
arXiv
(2022).
\doiurl{10.48550/ARXIV.2203.11258}.
\url{https://arxiv.org/abs/2203.11258}
\end{botherref}
\endbibitem

\bibitem{understanding-long-documents}
\begin{botherref}
\oauthor{\bsnm{Pham}, \binits{H.}},
\oauthor{\bsnm{Wang}, \binits{G.}},
\oauthor{\bsnm{Lu}, \binits{Y.}},
\oauthor{\bsnm{Florencio}, \binits{D.}},
\oauthor{\bsnm{Zhang}, \binits{C.}}:
Understanding Long Documents with Different Position-Aware Attentions.
arXiv
(2022).
\doiurl{10.48550/ARXIV.2208.08201}.
\url{https://arxiv.org/abs/2208.08201}
\end{botherref}
\endbibitem

\bibitem{grail-etal-2021-globalizing}
\begin{bchapter}
\bauthor{\bsnm{Grail}, \binits{Q.}},
\bauthor{\bsnm{Perez}, \binits{J.}},
\bauthor{\bsnm{Gaussier}, \binits{E.}}:
\bctitle{Globalizing {BERT}-based transformer architectures for long document
  summarization}.
In: \bbtitle{Proceedings of the 16th Conference of the European Chapter of the
  Association for Computational Linguistics: Main Volume},
pp. \bfpage{1792}--\blpage{1810}.
\bpublisher{Association for Computational Linguistics},
\blocation{Online}
(\byear{2021}).
\doiurl{10.18653/v1/2021.eacl-main.154}.
\burl{https://aclanthology.org/2021.eacl-main.154}
\end{bchapter}
\endbibitem

\bibitem{bertal}
\begin{botherref}
\oauthor{\bsnm{Zhang}, \binits{R.}},
\oauthor{\bsnm{Wei}, \binits{Z.}},
\oauthor{\bsnm{Shi}, \binits{Y.}},
\oauthor{\bsnm{Chen}, \binits{Y.}}:
{\{}BERT{\}}-{\{}AL{\}}: {\{}BERT{\}} for Arbitrarily Long Document
  Understanding
(2020).
\url{https://openreview.net/forum?id=SklnVAEFDB}
\end{botherref}
\endbibitem

\bibitem{beyond-512}
\begin{bchapter}
\bauthor{\bsnm{Yang}, \binits{L.}},
\bauthor{\bsnm{Zhang}, \binits{M.}},
\bauthor{\bsnm{Li}, \binits{C.}},
\bauthor{\bsnm{Bendersky}, \binits{M.}},
\bauthor{\bsnm{Najork}, \binits{M.}}:
\bctitle{Beyond 512 tokens: Siamese multi-depth transformer-based hierarchical
  encoder for long-form document matching}.
\bsertitle{CIKM '20},
pp. \bfpage{1725}--\blpage{1734}.
\bpublisher{Association for Computing Machinery},
\blocation{New York, NY, USA}
(\byear{2020}).
\doiurl{10.1145/3340531.3411908}.
\burl{https://doi.org/10.1145/3340531.3411908}
\end{bchapter}
\endbibitem

\bibitem{mahata2022ldkp}
\begin{botherref}
\oauthor{\bsnm{Mahata}, \binits{D.}},
\oauthor{\bsnm{Agarwal}, \binits{N.}},
\oauthor{\bsnm{Gautam}, \binits{D.}},
\oauthor{\bsnm{Kumar}, \binits{A.}},
\oauthor{\bsnm{Parekh}, \binits{S.}},
\oauthor{\bsnm{Singla}, \binits{Y.K.}},
\oauthor{\bsnm{Acharya}, \binits{A.}},
\oauthor{\bsnm{Shah}, \binits{R.R.}}:
Ldkp: A dataset for identifying keyphrases from long scientific documents.
arXiv preprint arXiv:2203.15349
(2022)
\end{botherref}
\endbibitem

\bibitem{query-based-kpe}
\begin{botherref}
\oauthor{\bsnm{Do{\v{c}}ekal}, \binits{M.}},
\oauthor{\bsnm{Smr{\v{z}}}, \binits{P.}}:
Query-based keyphrase extraction from long documents.
The International {FLAIRS} Conference Proceedings
\textbf{35}
(2022).
\doiurl{10.32473/flairs.v35i.130737}
\end{botherref}
\endbibitem

\bibitem{garg2022keyphrase}
\begin{botherref}
\oauthor{\bsnm{Garg}, \binits{K.}},
\oauthor{\bsnm{Chowdhury}, \binits{J.R.}},
\oauthor{\bsnm{Caragea}, \binits{C.}}:
Keyphrase Generation Beyond the Boundaries of Title and Abstract
(2022)
\end{botherref}
\endbibitem

\bibitem{kim-2010}
\begin{bchapter}
\bauthor{\bsnm{Kim}, \binits{S.N.}},
\bauthor{\bsnm{Medelyan}, \binits{O.}},
\bauthor{\bsnm{Kan}, \binits{M.-Y.}},
\bauthor{\bsnm{Baldwin}, \binits{T.}}:
\bctitle{Semeval-2010 task 5: Automatic keyphrase extraction from scientific
  articles}.
In: \bbtitle{Proceedings of the 5th International Workshop on Semantic
  Evaluation}.
\bsertitle{SemEval '10},
pp. \bfpage{21}--\blpage{26}.
\bpublisher{Association for Computational Linguistics},
\blocation{USA}
(\byear{2010})
\end{bchapter}
\endbibitem

\bibitem{duc2001}
\begin{bchapter}
\bauthor{\bsnm{Wan}, \binits{X.}},
\bauthor{\bsnm{Xiao}, \binits{J.}}:
\bctitle{Single document keyphrase extraction using neighborhood knowledge}.
In: \bbtitle{Proceedings of the 23rd National Conference on Artificial
  Intelligence - Volume 2}.
\bsertitle{AAAI'08},
pp. \bfpage{855}--\blpage{860}.
\bpublisher{AAAI Press}, \blocation{???}
(\byear{2008})
\end{bchapter}
\endbibitem

\bibitem{s2orc-corpus}
\begin{botherref}
\oauthor{\bsnm{Lo}, \binits{K.}},
\oauthor{\bsnm{Wang}, \binits{L.L.}},
\oauthor{\bsnm{Neumann}, \binits{M.}},
\oauthor{\bsnm{Kinney}, \binits{R.}},
\oauthor{\bsnm{Weld}, \binits{D.S.}}:
{GORC:} {A} large contextual citation graph of academic papers.
CoRR
\textbf{abs/1911.02782}
(2019)
{\href{https://arxiv.org/abs/1911.02782}{{arXiv:1911.02782}}}
\end{botherref}
\endbibitem

\bibitem{scibert}
\begin{bchapter}
\bauthor{\bsnm{Beltagy}, \binits{I.}},
\bauthor{\bsnm{Lo}, \binits{K.}},
\bauthor{\bsnm{Cohan}, \binits{A.}}:
\bctitle{{S}ci{BERT}: A pretrained language model for scientific text}.
In: \bbtitle{Proceedings of the 2019 Conference on Empirical Methods in Natural
  Language Processing and the 9th International Joint Conference on Natural
  Language Processing (EMNLP-IJCNLP)},
pp. \bfpage{3615}--\blpage{3620}.
\bpublisher{Association for Computational Linguistics},
\blocation{Hong Kong, China}
(\byear{2019}).
\doiurl{10.18653/v1/D19-1371}.
\burl{https://aclanthology.org/D19-1371}
\end{bchapter}
\endbibitem

\bibitem{openai2023gpt}
\begin{botherref}
\oauthor{\bsnm{OpenAI}}:
Gpt-4 technical report.
arXiv
(2023)
\end{botherref}
\endbibitem

\end{thebibliography}


\end{document}